\documentclass{article} 
\usepackage{iclr2023_conference,times}
\iclrfinalcopy

\usepackage{amsmath,amsfonts,bm}

\def\eqref#1{equation~\ref{#1}}

\def\1{\bm{1}}

\def\vp{{\bm{p}}}

\def\vx{{\bm{x}}}
\def\vy{{\bm{y}}}
\def\vz{{\bm{z}}}

\DeclareMathAlphabet{\mathsfit}{\encodingdefault}{\sfdefault}{m}{sl}
\SetMathAlphabet{\mathsfit}{bold}{\encodingdefault}{\sfdefault}{bx}{n}

\def\gN{{\mathcal{N}}}

\newcommand{\R}{\mathbb{R}}

\newcommand{\inputdim}{D}
\newcommand{\latentdim}{H}

\newcommand\idata{i\xspace}

\newcommand\dataix{^{(\idata)}}

\newcommand\nclass{C\xspace}

\usepackage{hyperref}
\usepackage{url}
\usepackage{subcaption}
\usepackage{adjustbox}
\usepackage{multirow}
\usepackage{makecell}
\usepackage{pgfplotstable}
\usepackage{booktabs}
\usepackage{xspace}
\usepackage{wrapfig}
\usepackage{cleveref}
\usepackage{colortbl}
\usepackage{bm}
\usepackage{tabularx}

        \definecolor{mycolor}{RGB}{220,220,220}
    
\title{Training, Architecture, and Prior \\for Deterministic Uncertainty Methods}

\author{%
  Bertrand Charpentier, Chenxiang Zhang, Stephan Günnemann \\
  Technical University of Munich, Germany\\
  \texttt{\{charpent,zch,guennemann\}@in.tum.de} \\
}

\begin{document}

\maketitle

\begin{abstract}
Accurate and efficient uncertainty estimation is crucial to build reliable Machine Learning (ML) models capable to provide calibrated uncertainty estimates, generalize and detect Out-Of-Distribution (OOD) datasets. To this end, Deterministic Uncertainty Methods (DUMs) is a promising model family capable to perform uncertainty estimation in a single forward pass. This work investigates important design choices in DUMs: 
\textbf{(1)} we show that \emph{training} schemes decoupling the core architecture and the uncertainty head schemes can significantly improve uncertainty performances. 
\textbf{(2)} we demonstrate that the core \emph{architecture} expressiveness is crucial for uncertainty performance and that additional architecture constraints to avoid feature collapse can deteriorate the trade-off between OOD generalization and detection. 
\textbf{(3)} Contrary to other Bayesian models, we show that the \emph{prior} defined by DUMs do not have a strong effect on the final performances.
\end{abstract}

\section{Introduction}
\label{sec:introduction}

\looseness=-1
Safety is critical to the adoption of deep learning in domains such as autonomous driving, medical diagnosis, or financial trading systems. A solution for this problem is to create reliable models capable to estimate the uncertainty of its own predictions. 
Different uncertainty types are divided in \textit{aleatoric} uncertainty quantified by the inherited noise in the data, thus irreducible; \textit{epistemic} uncertainty quantified by the modeling choice or lack of data, thus reducible; \textit{predictive} uncertainty, a combination of aleatoric and epistemic \citep{gal2016uncertainty}. 
In practice, high quality uncertainty estimates must be calibrated and able to detect Out-Of-Distribution (OOD) data like anomalies while preserving good Out-Of-Distribution (OOD) generalization performances like on dataset shifts.

Recently, a family of methods for uncertainty estimation named Deterministic Uncertainty Methods (DUMs) have emerged \citep{postels2022practicalitydum}. 
Contrary to uncertainty methods such as Ensembles \citep{lakshminarayanan2017ensembles}, MC Dropout \citep{gal2016dropout} or other Bayesian neural networks on weights \citep{blundell2015weight}, which require multiple forward passes to make predictions, DUMs only require a single forward pass, thus making them significantly more computationally efficient. 
Generally, DUMs are composed of three components with high potential impact on their performances: the \emph{training} procedure which is supposed to optimize the model toward high predictive and uncertainty performances,  the core \emph{architecture} which is supposed to define informative embeddings used to make predictions, and the \emph{prior} which is supposed to define the default uncertain predictions. In this work, we investigate the role of these three components on the quality of DUMs uncertainty estimates by evaluating calibration performances, OOD detection, and OOD generalization. Our main contributions are:
\vspace{-2mm}
\begin{itemize}
    \item \textbf{Training}: We show that \emph{decoupling the learning rates} of the core architecture and uncertainty heads of DUMs, \emph{jointly training} the core architecture and the uncertainty head of DUMs, and \emph{pretraining} with \emph{more data} and \emph{higher data quality} improve uncertainty performances. 
    \item \textbf{Architecture}: We demonstrate that the expressiveness of the core architecture defined by the \emph{architecture type}, \emph{architecture size}, and \emph{dimension of the latent space} is crucial for \emph{both} predictive and uncertainty performances. Further, we show that applying additional regularization constraints to avoid \emph{feature collapse} does not find better trade-off between OOD detection and generalization, even sometimes degrading performances.
    \item \textbf{Prior}: In contrast to Bayesian neural networks on weights where the choice of prior is critical \citep{wenzel2020prior_cpe, fortuin2022prior, noci2021prior_cpe, kapoor2022prior_cpe}, we empirically show that the choice of prior defined in the training loss or in the uncertainty head of DUMs has a relatively small effect on the final uncertainty performances.
\end{itemize}

\section{Deterministic Uncertainty Methods}
\label{sec:dum}

We consider a classification task where the goal is to predict the label $y\dataix \in \{ 1, \ldots , \nclass \}$ based on the input feature $\vx\dataix \in \R^{\inputdim}$. In this case, a DUM generally performs predictions in two main steps: \textbf{(1)} A core \emph{architecture} $f_\mathbf{\phi}$ maps the input features $\vx\dataix \in \R^{\inputdim}$ to a latent representation $\vz\dataix \in \R^{\latentdim}$, i.e. $f_\mathbf{\phi}(\vx\dataix)=\vz\dataix$. In practice, important design choices are the latent dimension $\latentdim$ and the architecture $f_\mathbf{\phi}$ which should be adapted to the task (see \cref{sec:architecture}). Further, multiple works proposed to apply additional bi-Lipschitz or reconstruction constraints to enrich the informativeness of the latent representation $\vz\dataix$ (see \cref{sec:architecture}). \textbf{(2)} An \emph{uncertainty head} $g_\mathbf{\psi}$ maps the latent representation $\vz\dataix$ to a predicted label $\hat{y}\dataix$ and an associated (aleatoric, epistemic, or predictive) uncertainty estimate $u\dataix$, i.e. $g_\mathbf{\psi}(\vz\dataix)=(\hat{y}\dataix, u\dataix)$. In practice, important design choices are the type of uncertainty head which are generally instantiated with a Gaussian Process (GP) \citep{liu2020sngp, van2021due, van2020uncertainty, bilos2019uncertainty, charpentier2022rl}, a density estimator \citep{charpentier2020pn, charpentier2022natpn, charpentier2022rl, stadler2021graph, bilos2019uncertainty, mukhoti2021ddu, postels2020mir, winkens2020contrastive, wu2020contrastive}, or an evidential model \citep{charpentier2020pn, charpentier2022natpn, charpentier2022rl, stadler2021graph, bilos2019uncertainty, malinin2018prior}, and the choice of the prior used by the uncertainty head (see \cref{sec:prior}). Beyond core architecture and uncertainty head, another important choice is the training procedure which can either couple or decouple the parameters of the core architecture and uncertainty head (see \cref{sec:training}).

In this work we focus on two recent DUMs which cover different types of uncertainty heads: Natural Posterior Network (NatPN) \cite{charpentier2022natpn} which learns an evidential distribution based on density estimation on the latent space, and Deterministic Uncertainty Estimator (DUE) \citep{van2021due} which learns a deep Gaussian Process by parametrizing learnable inducing points in the latent space (see \cref{appendix:dums} for further method details). While NatPN is capable to differentiate aleatoric, epistemic, and predictive uncertainty, DUE only outputs the predictive uncertainty. 
For all the experiments, we use the same default setup: we first \emph{pretrain} the encoder with the cross-entropy loss until convergence, then load the pretrained encoder and jointly \emph{train} the encoder and uncertainty head (see \cref{subsec:appendix_training,subsec:appendix_encoder,subsec:appendix_prior} for further compenent details). We report the predictive performance via accuracy, and the uncertainty performances with Brier Score and OOD detection results after averaging over 5 seeds (see \cref{sec:metric_details} for further metric details). We perform our experiments on MNIST \citep{lecun1998mnist}, CIFAR10 and CIFAR100 \citep{krizhevsky09cifar}, and Camelyon \citep{koh2021wilds}. OOD results reported in 
\cref{tab:training_schema,tab:training_pretrain,tab:encoder_architecture,tab:kernel} averages the uncertainty estimation from five OOD datasets: SVHN, STL10, CelebA, Camelyon and SVHN OODom \citep{netzer2011svhn, coates2011stl10, liu2015celeba}. Our code and additional material is available online\footnote{\url{https://www.cs.cit.tum.de/daml/training-architecture-prior-dum/}}.

\textbf{Related work.} Previous works survey OOD detection methods \citep{yang2021ooddet}, OOD generazilation methods \citep{shen2021oodgen}, or a wide range of uncertainty estimation methods \citep{gawlikowski2021survey,psaros2023survey,ulmer2021survey,abdar2021survey} by presenting key methods and challenges. These surveys do not focus on deterministic methods and do not make empirical analysis.
Other works propose great empirical studies to compare uncertainty estimation methods under shifts \citep{ovadia2019shift}, or analyze the role of the prior in Bayesian neural networks on weights \citep{wenzel2020prior_cpe, fortuin2022prior, noci2021prior_cpe, kapoor2022prior_cpe}. These works do not focus on DUMs. Closer to our work,  \citet{postels2022practicalitydum} compares methods in the DUMs family and demonstrate calibration limitations. In contrast, we evaluate the role of \emph{components} in DUMs and show that carefully specifying \emph{training}, \emph{architecture}, or \emph{prior} can improve uncertainty metrics like calibration and OOD detection but also ID and OOD predictive performances.


\section{Training for DUMs}
\label{sec:training}

In this section, we study the importance of the training procedure in the performance of DUMs. To this end, we look at \emph{decoupling the learning rates} of the core encoder architecture and the uncertainty head, different \emph{training schemes}, and different \emph{pretraining schemes}.

\begin{figure}[!htb]
    \centering
    \includegraphics[width=0.8\linewidth]{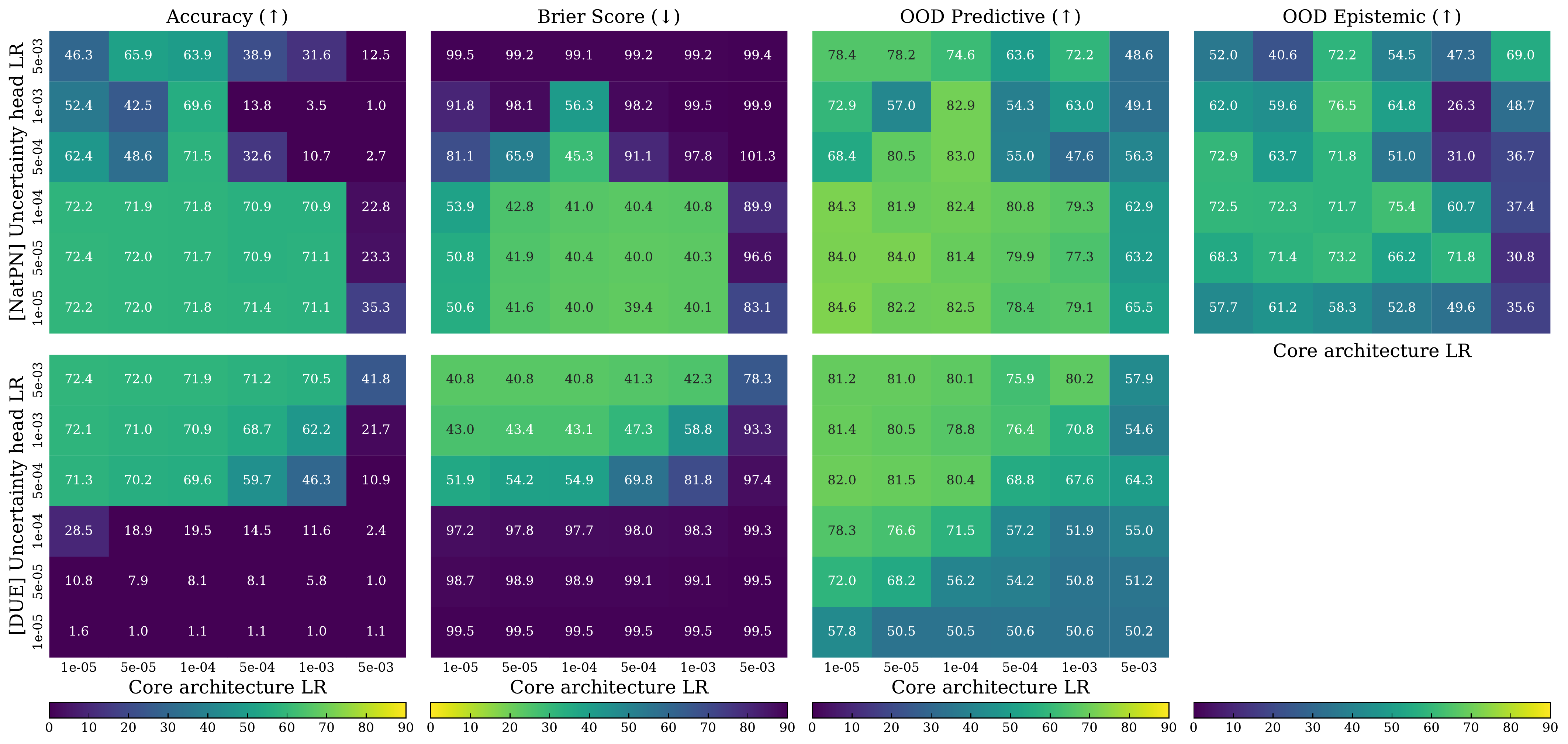}
    \caption{Results of DUMS on CIFAR100 with ResNet18 when \textbf{decoupling learning rates} of the core architecture and the uncertainty head. Decoupling learning rates improve DUMs performance. }
    \label{fig:decoupled_test}
\end{figure}

\textbf{Decoupling learning rates.} We decouple the learning rates of the core architecture and the uncertainty head. We show the validation results for CIFAR100 as ID and SVHN as OOD with the core architecture ResNet18 in \cref{fig:decoupled_test}.
\underline{\textit{Observation:}} We observe that, when using different learning rates for the core architecture and the uncertainty head, NatPN improves Brier Score and OOD epistemic results and DUE significantly improves both predictive and uncertainty results. Hence, this shows that decoupling learning rates can improve results of DUMs, thus suggesting that the core architecture and the uncertainty head have training dynamics which requires different considerations.

\textbf{Training schemes.} We compare two settings: the \textit{joint} training in which we jointly train the weights of the core architecture and uncertainty head, and the \textit{sequential} training in which we only train the uncertainty head by keeping the weights of the pretrained core architecture fixed. For each of the setting, we apply two additional techniques to stabilize the training: adding a \textit{batch normalization} to the last layer of the encoder to enforce latent representations to locate in a normalized region \citep{ioffe2015bn,charpentier2022natpn}, and \textit{resetting the last layer} to retrain its weights to improve robustness to spurious correlation \citep{kirichenko2022reset}. We show the results for CIFAR100 as ID and five difference OOD datasets with the ResNet18 as core architecture in \cref{tab:training_schema} and additional results in the appendix \cref{tab:training_schema_ood}.
\underline{\textit{Observation:}} We observe that, compared to its sequential counterpart, joint training consistently improves DUMs performance for most metrics, thus suggesting that joint training should be preferred in practice for DUMs. Furthermore, while the GP-based method DUE does not benefit from stabilization techniques, we observe that they can significantly increase performance of the density-based method NatPN. This behavior is intuitively explained by the practical difficulty to accurately fit densities in high dimensional latent space. This can be significantly improved by using more powerful density estimator (see \cref{tab:normalizing_flow} in appendix).

\textbf{Pretraining schemes.} We compare multiple training schemes which differ in terms of \emph{amount} and \emph{quality} of data used for pretraining. Hence, we do not pretrain the core architecture or pretrain it with 10\% of CIFAR100, 100\% of CIFAR100 without and with Gaussian noise, or ImageNet. We show the results for CIFAR100 as ID and five different OOD datasets with ResNet50 as core architecture in \cref{tab:training_pretrain} and additional results in the appendix \cref{tab:training_pretrain_ood}.
\underline{\textit{Observation:}} We observe that, while too few data for pretraining does not improve final performance of DUMs, the overall performance significantly increase when the encoder is pretrained with high quantity and high quality of data.  Similarly to \citet{kirichenko2020why}, this suggests that the embedding quality is important to improve uncertainty quantification. Here, we show additionally that embeddings pretrained with many high quality data are crucial to facilitate the prediction of the uncertainty head.

\begin{table}[!htb]
\centering
\tiny

\begin{minipage}[t]{0.48\textwidth}
\caption{Results of DUMs on CIFAR100 with ResNet18 under different \textbf{training schemes} using \textit{joint/sequential} training with no additional layer, an additional batch norm layer, or resetting the last layer. Gray cells indicate the best between \textit{joint/sequential} while bold numbers indicate the best overall. OOD results are averaged over OOD datasets. We observe that joint training works best and stabilization techniques can improve performances. }
\label{tab:training_schema}

\resizebox{\textwidth}{!}{%
\begin{tabular}{llccccc}
\toprule
\textbf{Method} &\textbf{Train Schema} &\textbf{Accuracy ($\uparrow$)} &\textbf{Brier Score ($\downarrow$)} &\textbf{OOD Pred. ($\uparrow$)} &\textbf{OOD Epis. ($\uparrow$)} \\
\midrule
\multirow{6}{*}{NatPN} 
& joint &71.12 $\pm$ 0.18 & \cellcolor{mycolor}41.06 $\pm$ 0.18 & \cellcolor{mycolor}75.17 $\pm$ 1.60 & \cellcolor{mycolor}63.94 $\pm$ 2.80 \\
& joint + bn &71.60 $\pm$ 0.14 &\cellcolor{mycolor}41.11 $\pm$ 0.12 & 74.22 $\pm$ 0.94 & \cellcolor{mycolor}66.17 $\pm$ 2.55 \\
& joint + reset &71.61 $\pm$ 0.18 & \cellcolor{mycolor}\textbf{40.76 $\pm$ 0.18} &\cellcolor{mycolor}\textbf{75.35 $\pm$ 0.71} & \cellcolor{mycolor}\textbf{69.02 $\pm$ 1.49} \\
& sequential & \cellcolor{mycolor}\textbf{72.00 $\pm$ 0.19} &42.20 $\pm$ 0.09 &75.09 $\pm$ 0.86 &53.49 $\pm$ 2.56 \\
& sequential + bn & \cellcolor{mycolor}71.98 $\pm$ 0.18 & 42.39 $\pm$ 0.11 & \cellcolor{mycolor}75.01 $\pm$ 0.86 &52.34 $\pm$ 2.81 \\
& sequential + reset & \cellcolor{mycolor}71.79 $\pm$ 0.17 &40.95 $\pm$ 0.14 &74.63 $\pm$ 0.85 &61.90 $\pm$ 2.14 \\
\midrule
\multirow{6}{*}{DUE} 
& joint & \cellcolor{mycolor}\textbf{72.33 $\pm$ 0.11} & \cellcolor{mycolor}\textbf{40.80 $\pm$ 0.11} &74.74 $\pm$ 0.89 &- \\
& joint + bn & \cellcolor{mycolor}72.30 $\pm$ 0.09 & \cellcolor{mycolor}40.85 $\pm$ 0.12 &74.63 $\pm$ 0.95 &- \\
& joint + reset & \cellcolor{mycolor}71.94 $\pm$ 0.12 & \cellcolor{mycolor}41.43 $\pm$ 0.12 &74.89 $\pm$ 0.76 &- \\
& sequential &72.07 $\pm$ 0.10 &41.66 $\pm$ 0.10 & \cellcolor{mycolor}74.82 $\pm$ 0.90 &- \\
& sequential + bn &72.04 $\pm$ 0.13 &41.73 $\pm$ 0.11 & \cellcolor{mycolor}74.88 $\pm$ 0.95 &- \\
& sequential + reset &71.56 $\pm$ 0.14 &42.30 $\pm$ 0.11 & \cellcolor{mycolor}\textbf{75.08 $\pm$ 1.01} &- \\
\bottomrule
\end{tabular}
}\end{minipage}\quad%
\begin{minipage}[t]{0.48\textwidth}
\caption{Results of DUMs with ResNet50 under different \textbf{pretraining schemes} using no pretraining, pretraining on 10\% of CIFAR100, 100\% of CIFAR100 without Gaussian noise and with Gaussian noise, or ImageNet. OOD results are averaged over OOD datasets. Bold numbers indicate best results among all settings. We observe that high quantity and high quality of data can improve performances.}
\label{tab:training_pretrain}

\resizebox{\textwidth}{!}{%
\begin{tabular}{llccccc}\toprule
\textbf{Method} &\textbf{Pretrain Schema} &\textbf{Accuracy ($\uparrow$)} &\textbf{Brier Score ($\downarrow$)} &\textbf{OOD Pred. ($\uparrow$)} &\textbf{OOD Epis. ($\uparrow$)} \\
\midrule
\multirow{6}{*}{NatPN} &None &78.45 $\pm$ 1.94 &30.16 $\pm$ 2.57 &79.85 $\pm$ 3.30 &85.81 $\pm$ 2.05 \\
&C100 (10\%) &67.25 $\pm$ 0.71 &44.69 $\pm$ 0.94 &68.10 $\pm$ 1.90 &78.50 $\pm$ 3.27 \\
&C100 (100\%) + $\gN(0.5)$ &72.50 $\pm$ 1.07 &39.36 $\pm$ 1.43 &68.66 $\pm$ 3.76 &73.54 $\pm$ 1.84 \\
&C100 (100\%) + $\gN(0.1)$ &75.25 $\pm$ 0.61 &35.99 $\pm$ 0.88 &69.78 $\pm$ 4.65 &75.81 $\pm$ 2.85 \\
&C100 (100\%) &76.31 $\pm$ 0.45 &34.32 $\pm$ 0.51 &78.95 $\pm$ 3.19 &76.91 $\pm$ 2.64 \\
&ImageNet &\textbf{84.22 $\pm$ 0.12} &\textbf{23.67 $\pm$ 0.27} &\textbf{84.95 $\pm$ 1.48} &\textbf{89.08 $\pm$ 0.70} \\
\midrule
\multirow{6}{*}{DUE} &None &72.41 $\pm$ 0.24 &47.35 $\pm$ 0.25 &80.04 $\pm$ 1.28 &- \\
&C100 (10\%) &63.86 $\pm$ 0.58 &50.94 $\pm$ 0.53 &72.44 $\pm$ 1.32 &- \\
&C100 (100\%) &76.38 $\pm$ 0.35 &36.89 $\pm$ 0.50 &81.71 $\pm$ 1.87 &- \\
&C100 (100\%) + $\gN(0.5)$ &72.10 $\pm$ 1.00 &42.48 $\pm$ 1.18 &74.89 $\pm$ 1.97 &- \\
&C100 (100\%) + $\gN(0.1)$ &75.31 $\pm$ 0.91 &38.31 $\pm$ 1.22 &79.43 $\pm$ 1.93 &- \\
&ImageNet &\textbf{82.42 $\pm$ 0.14} &\textbf{28.09 $\pm$ 0.19} &\textbf{90.24 $\pm$ 0.51} &- \\
\bottomrule
\end{tabular}
}\end{minipage}
\end{table}

\section{Architecture for DUMs}
\label{sec:architecture}

In this section, we study the impact of the architecture component in DUMs. To this end, we look at different \emph{latent dimensions}, different \emph{architectural types and size}, and applying different \emph{regularization constraints} to avoid \textit{feature collapse} \citep{van2021due}. 

\textbf{Latent dimension.} We vary the dimension of the output space of the core architecture. We show the results for each pair of ID dataset and its distribution shifted OOD dataset (MNIST/CMNIST, CIFAR/CIFAR-C, CamelyonID/CamelyonOOD) with the core architecture ResNet18 for MNIST/CIFAR, and WideResNet-28-10 for Camelyon in \cref{fig:latent} and additional uncertainty estimation results in the appendix \cref{fig:latent_ood}. \underline{\textit{Observation:}} We observe that increasing the latent dimensions leads to improvement for DUMs on ID and OOD datasets with particularly significant improvement for NatPN (see \cref{fig:latent}). This suggests that higher latent dimensions are more expressive by encoding more information. However, we observe that  a too high latent dimension can degrade OOD detection performance by causing numerical instabilities in the training (see \cref{fig:latent_ood}), suggesting a trade-off between OOD generalization and OOD detection.

\begin{figure}[!htb]
    \centering
    \includegraphics[width=0.8\textwidth]{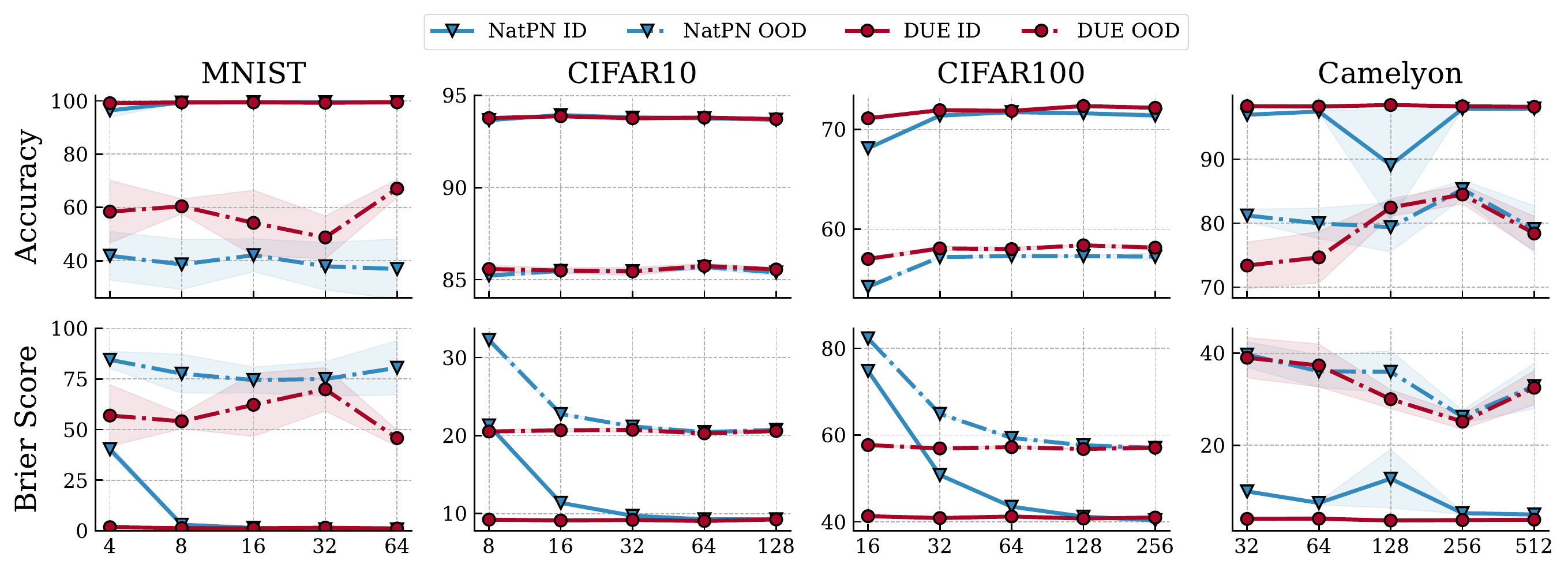}
    \caption{Results of DUMs when varying the \textbf{latent dimension size}. We observe that increasing the latent dimension consistently leads to similar or better predictive performance.}
    \label{fig:latent}
\end{figure}

\textbf{Architecture type and size.} We compare the influence of the type and size of the core architecture on the performance of DUMs. We consider residual, convolutional, and transformer architectures like ResNet18, ResNet50, EfficientNetV2, and Swin \citep{he2016resnet, tan2021effcientnet, liu2021swin}. We show the results for DUMs trained on CIFAR100 as ID with the different core architectures in \cref{tab:encoder_architecture} and additional results at appendix \cref{tab:encoder_architecture_ood}. \underline{\textit{Observation:}} We observe that models with more parameters achieve better results. In particular, ResNet50 achieves significantly better results than ResNet18. Further, more recent core architectures like EfficientNetV2 and Swin are better calibrated and more expressive leading to a better overall performance. This can be explained by the fact that they are more expressive and provide more informative embeddings for the uncertainty head to operate on. This aligns with \citet{minderer2021calibration} which states that the architecture type is important for the calibration properties. 


\begin{table}[!htp]\centering
\caption{Results of DUMs for different \textbf{architecture types.} including residual, convolutional, and transformer architectures on CIFAR100.  OOD results are averaged over OOD datasets. Bold numbers indicate best results among all settings. Larger and more recent architectures are better calibrated with similar or better uncertainty estimation.}
\label{tab:encoder_architecture}
\tiny

\resizebox{0.6\textwidth}{!}{%
\begin{tabular}{llcccccc}\toprule
\textbf{Method} &\textbf{Architecture} &\textbf{\#Parameters} &\textbf{Accuracy ($\uparrow$)} &\textbf{Brier Score ($\downarrow$)} &\textbf{OOD Pred. ($\uparrow$)} &\textbf{OOD Epis. ($\uparrow$)} \\
\midrule
\multirow{4}{*}{NatPN} &ResNet18 &11.6M &80.31 $\pm$ 0.09 &33.69 $\pm$ 0.15 &87.49 $\pm$ 1.90 &81.79 $\pm$ 1.38 \\
&ResNet50 &25.5M &84.22 $\pm$ 0.12 &23.67 $\pm$ 0.27 &84.95 $\pm$ 1.48 &89.08 $\pm$ 0.70 \\
&EffNet\_V2\_S &21.4M &\textbf{88.43 $\pm$ 0.10} &\textbf{17.08 $\pm$ 0.10} &\textbf{87.79 $\pm$ 0.77} &89.47 $\pm$ 0.59 \\
&Swin\_T &28.2M &87.99 $\pm$ 0.09 &18.48 $\pm$ 0.06 &85.91 $\pm$ 1.17 &\textbf{90.23 $\pm$ 0.81} \\
\midrule
\multirow{4}{*}{DUE} &ResNet18 &11.6M &78.85 $\pm$ 0.19 &36.57 $\pm$ 0.15 &88.04 $\pm$ 0.67 &- \\
&ResNet50 &25.5M &82.42 $\pm$ 0.14 &28.09 $\pm$ 0.19 &\textbf{90.24 $\pm$ 0.51} &- \\
&EffNet\_V2\_S &21.4M &86.92 $\pm$ 0.08 &\textbf{21.07 $\pm$ 0.06} &89.43 $\pm$ 0.67 &- \\
&Swin\_T &28.2M &\textbf{86.93 $\pm$ 0.05} &23.23 $\pm$ 0.05 &89.90 $\pm$ 0.36 &- \\
\bottomrule
\end{tabular}
}
\end{table}

\begin{figure}
\begin{minipage}[t]{0.48\textwidth}
    \centering
    \includegraphics[width=0.9\textwidth]{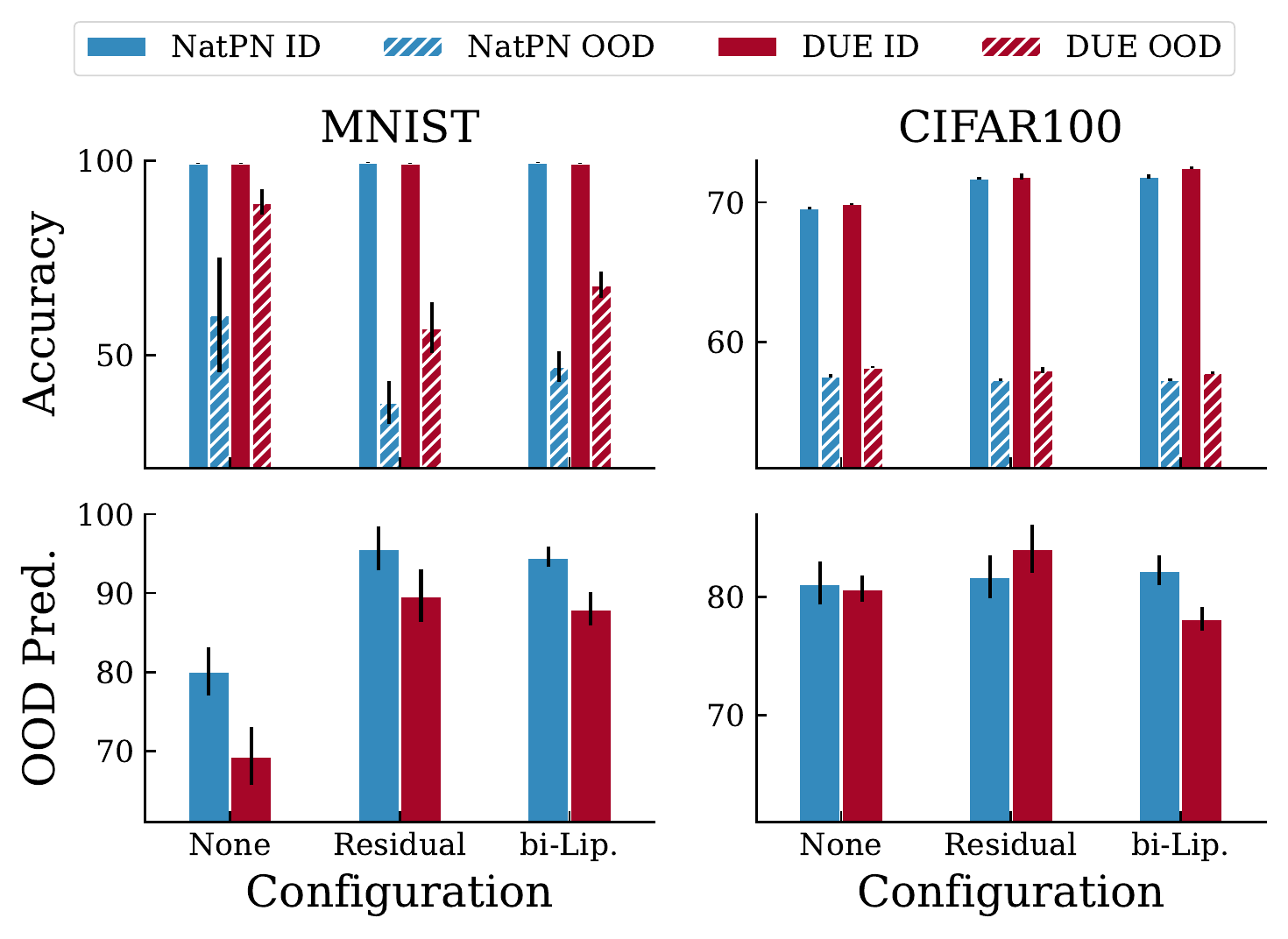}
    \caption{Results OOD generalization and detection of DUMs with none, residual and bi-lipschitz \textbf{architecture constraints} on MNIST/CMNIST and CIFAR/CIFAR-C. Bi-lipschitz can improve OOD detection by mitigating feature collapse (see \cref{fig:bi_toy}) at the expense of degrading OOD generalization.}
    \label{fig:bi}
\end{minipage}\quad%
\begin{minipage}[t]{0.48\textwidth}
    \centering
    \includegraphics[width=0.9\textwidth]{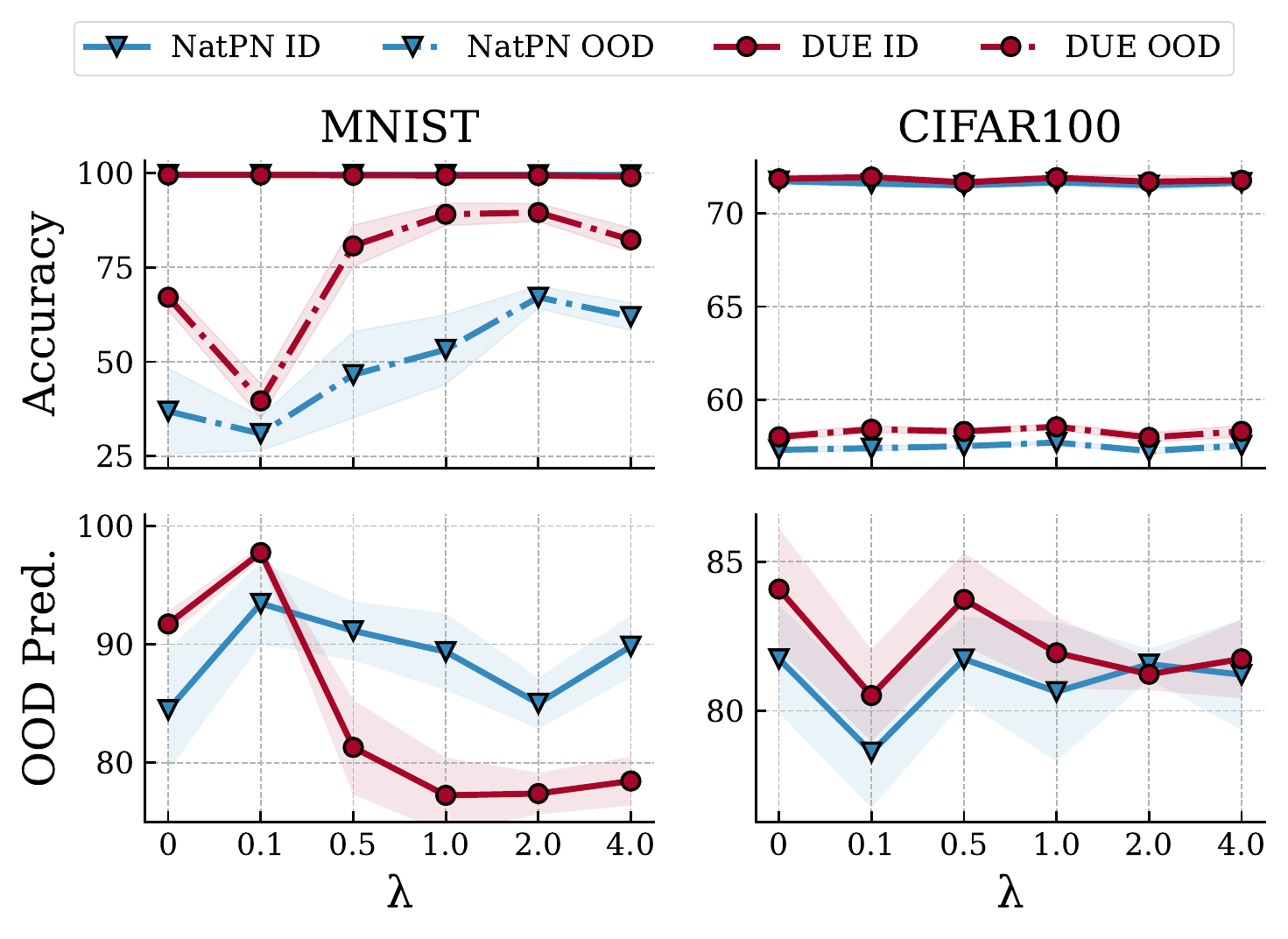}
    \caption{Results OOD generalization and detection of DUMs with reconstruction \textbf{architecture constraints} on MNIST/CMNIST and CIFAR/CIFAR-C. Increasing the reconstruction strength $\lambda$ improves the OOD generalization on simple MNIST/CMNIST dataset but fails for complex datasets. Reconstruction fails to improve OOD detection since it does not avoid feature collapse (see \cref{fig:rec_toy}). }
    \label{fig:rec}
\end{minipage}
\end{figure}

\textbf{Regularization constraints.} \emph{Feature collapse} is a phenomenon where a model may discard important parts of the input information during its training phase, which may degrade OOD detection performance \citep{van2021due}. Two techniques to avoid feature collapses are \textit{bi-Lipschitz} constraints via combining residual connections and lipschitz constraints \citep{liu2020sngp}, and \textit{reconstruction} constraints via adding an additional reconstruction term in the loss \citep{postels2020reconstruction}. We show the results for DUMs trained on the datasets MNIST and CIFAR100 with ResNet18 in \cref{fig:bi,fig:rec} and additional results for other datasets (Toy dataset, CIFAR10, Camelyon) at \cref{subsec:appendix_encoder}. \underline{\textit{Observation:}} We observe that the reconstruction technique is not capable to avoid feature collapse. Indeed, we show  that, even with reconstruction constraints, some (non-discriminative) features can completely collapse (see \cref{fig:rec_toy_collapse,fig:rec_toy} for toy examples). Hence, while this method can lead to small OOD improvements on simple tasks (see e.g. MNIST in \cref{fig:rec}), this benefit does not generalize to more complex tasks (see e.g. CIFAR100 in \cref{fig:rec}). In contrast, we observe that bilipschitz constraints indeed mitigate the collapse of features (see \cref{fig:bi_toy_collapse,fig:bi_toy} for toy examples), leading to similar or higher OOD detection performance (see \cref{fig:bi}). The mitigation of feature collapse can be mostly assigned to the residual connection constraints. However, bilipschitz constraints can improve OOD detection results on simple tasks (e.g. MNIST, CIFAR10), it degrades OOD generalization performance (see \cref{fig:bi}) and does not significantly improve OOD detection on more complex tasks (see e.g. Camelyon, CIFAR100 in \cref{fig:bi_bar_full}). Intuitively, maintaining features which are not discriminative to the task might introduce spurious correlations, thus degrading performances. E.g. enforces the architecture to encode the color feature in the latent space decreases the performance of the OOD CMNIST datasets after training on the ID MNIST dataset. 

\section{Prior for DUMs}
\label{sec:prior}

In this section, we study the effect that the prior component has in DUMs. More specifically, we investigate the relationships between aleatoric uncertainty and the prior specified for DUMs. In particular, this is motivated by \citet{kapoor2022prior_cpe} which shows that using priors that forces model to be confident on the training data points can improve its performance by explicitly accounting for aleatoric uncertainty.
To this end, we look at \textit{entropy regularization} defining a training prior in the loss, \textit{prior evidence} and \textit{kernel function} defining a functional prior in the uncertainty head.

\begin{figure}[!htb]
    \centering
    \includegraphics[width=\textwidth]{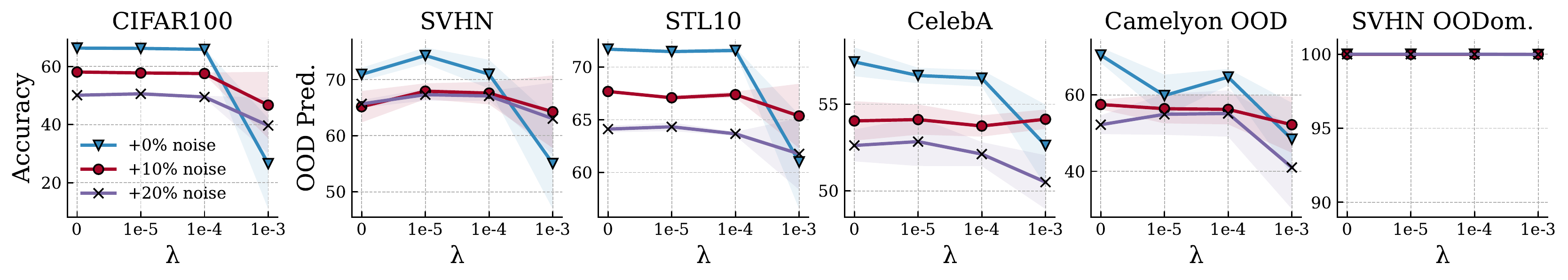}
    \includegraphics[width=\textwidth]{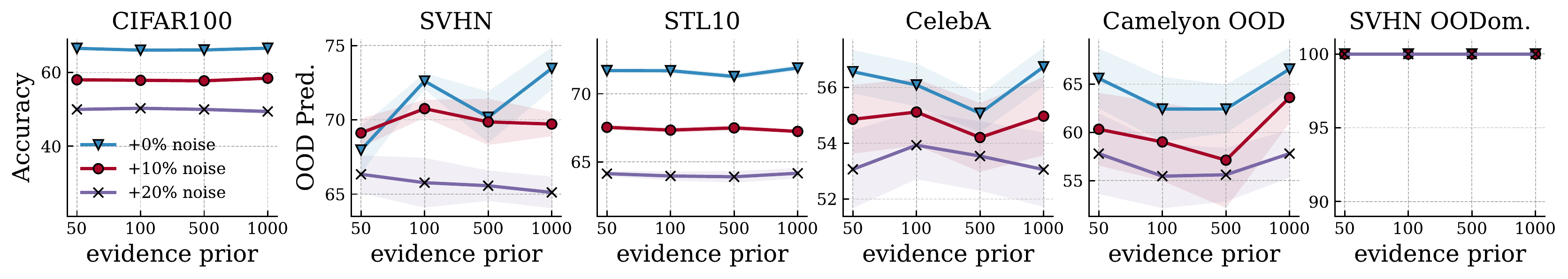}
    
    \caption{Results of enforcing different \textbf{prior} in NatPN on CIFAR100 by changing the (top) \textit{entropy regularization} $\lambda$ and the (bottom) \textit{evidence prior} $n^{prior}$. Different priors do not lead consistent results improvements.}
    \label{fig:prior_ood}
\end{figure}

\textbf{Prior.} We compare different prior specifications including \textit{entropy regularization} defining a training prior in the loss, \textit{prior evidence} and \textit{kernel function} defining a functional prior in the uncertainty head. Entropy regularization is the entropy term $H(Q)$ in the Bayesian loss used to train NatPN which encourages a (uniform) prior distributions with high entropy \citep{charpentier2022natpn}. We control the strength of the regularization factor $\lambda$. Further, NatPN also explicitly defines a prior via the parameters $\chi^{prior}$ and $n^{prior}$. While  $\chi^{prior}$ defines the default categorical prediction via a uniform categorical distribution, the evidence parameter $n^{prior}$ defines the prior number of pseudo-observations and can be varied. Finally, we vary the prior of DUE by using different kernel functions in the learned GP including Matern kernel, RQ kernel, and RBF \citep{rasmussen2006gp}. \underline{\textit{Observation:}} Contrary to other Bayesian neural networks \citep{kapoor2022prior_cpe}, we observe that predictive and uncertainty performances of DUMs are not very sensitive to the prior specification (see \cref{fig:prior_ood,fig:prior_ood_brier,tab:kernel}), thus suggesting a higher robustness to prior mispecification. Nonetheless, a too strong entropy regularization toward an uniform prior degrades more performance of DUMs trained on dataset with low label noise than on high label noise. This suggests that a too high discrepancy between the model prior and the dataset aleatoric uncertainties can impact performance.

 \vspace{-3mm}
\begin{table}[!htp]\centering
\caption{Results of enforcing different \textbf{prior} in DUE on CIFAR100 and Camelyon by changing the kernel function. OOD results are averaged over OOD datasets. Different priors lead to similar performance. }
\label{tab:kernel}
\vspace{-3mm}
\tiny
\resizebox{0.6\textwidth}{!}{%
\begin{tabular}{lcccccc}\toprule
&\multicolumn{3}{c}{\textbf{CIFAR100}} &\multicolumn{3}{c}{\textbf{Camelyon}} \\
\cmidrule(lr){2-4}\cmidrule(lr){5-7}
\textbf{Kernel} &\textbf{Accuracy ($\uparrow$)} &\textbf{Brier Score ($\downarrow$)} &\textbf{OOD Pred. ($\uparrow$)} &\textbf{Accuracy ($\uparrow$)} &\textbf{Brier Score ($\downarrow$)} &\textbf{OOD Pred. ($\uparrow$)} \\
\midrule
Matern52 &71.80 $\pm$ 0.18 &41.37 $\pm$ 0.24 &75.90 $\pm$ 1.18 &79.81 $\pm$ 2.72 &32.46 $\pm$ 3.22 &58.86 $\pm$ 6.20 \\
Matern32 &71.80 $\pm$ 0.21 &41.62 $\pm$ 0.22 &\textbf{76.15 $\pm$ 1.18} &80.23 $\pm$ 2.71 &32.77 $\pm$ 3.20 &58.50 $\pm$ 5.83 \\
Matern12 &71.70 $\pm$ 0.18 &43.10 $\pm$ 0.22 &75.70 $\pm$ 1.19 &79.30 $\pm$ 2.96 &32.67 $\pm$ 3.28 &\textbf{59.13 $\pm$ 6.39} \\
RQ &71.83 $\pm$ 0.19 &\textbf{41.16 $\pm$ 0.25} &75.93 $\pm$ 1.21 &80.31 $\pm$ 2.55 &32.22 $\pm$ 3.14 &58.69 $\pm$ 6.09 \\
RBF &\textbf{71.85 $\pm$ 0.19} &41.17 $\pm$ 0.24 &76.14 $\pm$ 1.19 &\textbf{80.45 $\pm$ 2.49} &\textbf{32.13 $\pm$ 3.11} &58.86 $\pm$ 5.91 \\
\bottomrule
\end{tabular}
}
\vspace{-3mm}
\end{table}

\section{Conclusion}
\label{sec:conclusion}

We investigate important design choice in DUMs. We show that \emph{training} of DUMs can be improved by decoupling the the optimization of the core architecture and the uncertainty head. We show that expressive core \emph{architecture} can improve DUMs performances. In contrast, additional constraints to avoid feature collapse do not consistently lead to better performance, potentially degrading the OOD generalization and detection trade-off. Finally, we show that the choice of \emph{prior} for DUMs does not lead to important performance improvements.

\newpage
\bibliography{iclr2023_conference}
\bibliographystyle{iclr2023_conference}

\newpage
\appendix
\section{Appendix}

\subsection{Deterministic Uncertainty Methods}
\label{appendix:dums}

\textbf{NatPN}. The deep Bayesian uncertainty model NatPN \citep{charpentier2022natpn} can be decomposed into these steps: 
\textbf{(1)} a core architecture predicts one latent representation of the input $\vx^{(i)}$ i.e. \smash{$\vz^{(i)} = f_{\bm{\phi}}(\vx^{(i)}) \in \mathbb{R}^H$}, 
\textbf{(2)} While a density estimator \smash{$\mathbb{P}(.|\bm{w})$} predicts the evidence parameter update \smash{$n^{(i)}=N_H\mathbb{P}(z^{(i)}|\bm{w})$} where the $N_H$ is a scaling factor named certainty budget, a single linear decoder $g_{\bm{\psi}}$ outputs the parameter update \smash{$\boldsymbol{\chi}^{(i)} = g_{\bm{\psi}}(\vz^{(i)}) \in \mathbb{R}^L$}, which can be viewed as a softmax output prediction.
\textbf{(3)} We perform an input-dependent Bayesian update which can be expressed in a closed-form as:
$$\mathbb{Q}(\boldsymbol{\theta}^{(i)}|\boldsymbol{\chi}^{\text{post},(i)}, n^{\text{post},(i)}) = \text{exp}(n^{\text{post},(i)}\boldsymbol{\theta}^{(i) T}\boldsymbol{\chi}^{\text{post},(i)}-n^{\text{post},(i)}A(\boldsymbol{\theta}^{(i)}))$$
$$\text{where} \qquad \boldsymbol{\chi}^{\text{post},(i)}=\frac{n^{\text{prior}}\boldsymbol{\chi}^{\text{prior}}+n^{(i)}\boldsymbol{\chi}^{(i)}}{n^{\text{prior}}+n^{(i)}}, \quad n^{\text{post},(i)}=n^\text{prior}+n^{(i)}$$
where $\boldsymbol{\chi}^\text{prior}, n^\text{prior}$ are fixed prior parameters, and  $\boldsymbol{\chi}^{\text{post}, (i)}, n^{\text{post}, (i)}$ are the input-dependent posterior parameters.
For the classification, the variable $\boldsymbol{\theta}^{(i)}$ represents the normalized categorical vector $\vp^{(i)}$. The predictive uncertainty is computed via the entropy of the predictive categorical distribution, and the epistemic uncertainty is computed via the evidence parameter $n^{post,(i)}$. We train all the components of neural network parameters $\{\bm{\phi}, \bm{w}, \bm{\psi}\}$ jointly with the Bayesian loss \citep{charpentier2022natpn}:
\begin{equation}
    \begin{split}
        \mathcal{L}^{(i)} \propto \mathbb{E}[\boldsymbol{\theta}^{(i)}]^T u(y^{(i)})-\mathbb{E}[A(\boldsymbol{\theta}^{(i)})]- \lambda\mathbb{H}[\mathbb{Q}^{post,(i)}] 
    \end{split}
\end{equation}
where $\lambda$ is the regularization factor of the entropy term representing. We refer to \citep{charpentier2022natpn} for a more detailed description of the method.

\textbf{DUE.} The deep kernel learning method DUE \citep{van2021due} can be decomposed into these steps: \textbf{(1)} a core architecture predicts one latent representation of the input $\vx^{(i)}$ i.e. \smash{$\vz^{(i)} = f_{\bm{\theta}}(\vx^{(i)}) \in \mathbb{R}^H$}, 
\textbf{(2)} a Gaussian Process defined from a fixed set of $K$ learnable inducing points \smash{$\{\bm{\phi}_{k}\}_{k=1}^{K}$} and a predefined positive definite kernel \smash{$\kappa(\cdot, \cdot)$} predicts the mean \smash{$\mu(\vx^{(i)})$} and the variance \smash{$\sigma(\vx^{(i)})$} of a Gaussian distribution, and 
\textbf{(3)} we apply softmax to the mean output \smash{$\mu(\vx^{(i)})$} for the classification prediction, i.e. $\vp^{(i)} = \text{softmax}(\mu(\vx^{(i)})$. We train the neural network parameters $\bm{\theta}$ and the inducing points \smash{$\{\bm{\phi}_{k}\}_{k=1}^{K}$} jointly with a variational ELBO loss. 
For classification, the predictive uncertainty is computed as the entropy of the predictive categorical distribution. 
We refer to \citep{van2021due} for a more detailed description of the method.




\subsection{Dataset details}
\label{appendix:dataset}
 We split all the training datasets into train, validation and test sets. For all the datasets, the test set is fixed while the training/validation sets are split in 80/20\% respectively. The random split of training/validation sets change depending on the seeds to ensure more diversity.  
 
\textbf{MNIST} \citep{lecun1998mnist}. Image classification dataset. Similarly as in \cite{arjovsky2019cmnist}, we create the CMNIST dataset for domain generalization experiments by expanding the input's size to 3 x 28 x 28 and zeroing one of the three channels. For OOD detection we use the test set of MNIST as ID dataset and compare to: KMNIST \citep{clanuwat2018kmnist}, CIFAR10, CMNIST, and KMNIST OODom, where we scale the input by 255. The batch size used is 512.

\textbf{CIFAR} \citep{krizhevsky09cifar}. Image classification dataset. We apply two data augmentations methods to the training data:the random horizontal flip and random cropping with padding equal to 4. For domain generalization we use the corrupted version CIFAR-C \citep{hendrycks2019corrupted} and report the average metric of 15 corruptions for the level of corruption severity of 1. For OOD detection we use the test set of CIFAR10 as ID dataset and compare to: SVHN \citep{netzer2011svhn}, STL10 \citep{coates2011stl10}, CelebA \citep{liu2015celeba}, Camelyon (Test OOD), and SVHN OODom. Since the Camelyon (Test OOD) dataset is large (85k), we extract only 10k subset of images as the OOD datasets. The batch size used is 128.

\textbf{Camelyon} \citep{koh2021wilds}. Image classification dataset. We apply two data augmentations emthods  to the training data: random horizontal flip and random rotation of 15 degrees. For domain generalization the dataset already provide the distribution shifted validation and test splits. For OOD detection we use the ID validation set of Camelyon as ID dataset (the ID  test set is not available) and compare to: SVHN, STL10, CelebA, Camelyon (Test OOD), and SVHN OODom. The batch size used is 32.

Each OOD dataset is rescaled to the same size as the ID dataset and normalized with zero mean and unit variance based on the statistics of ID dataset (for the Camelyon dataset we don't apply any normalization as in \cite{koh2021wilds}). 
\subsection{Metric details}
\label{sec:metric_details}


\textbf{Accuracy.} The standard accuracy $\frac{1}{N}\sum_i\1[y^{*,(i)} = y^{(i)}]$ is used, where $y^{*,(i)}$ is the true label and $y^{(i)}$ is the predicted label. 

\textbf{Calibration.} The Brier score $\frac{1}{C}\sum_i^N||\vp^{(i)} - \vy^{*, (i)}||$ is used, where $\vp^{(i)}$ is the predicted softmax probability and $\vy^{*, (i)}$ is the one-hot encoded true label. Lower calibration indicates a better calibrated model. Note that in constrast with the Expected Calibration Error (ECE), the Brier score is a strictly proper scoring rule which makes it a particularly good evaluation metric for calibration \cite{gneiting2007proper}.

\textbf{OOD Generalization.} We apply accuracy and calibration to the distribution shifted OOD dataset and compare the results with the ID dataset to estimate the model's ability for generalization.

\textbf{OOD Detection.} We treat this task as a binary classification, where we assign class 1 to ID data and class 0 to OOD data using the aleatoric, epistemic, and predictive uncertainty estimates as scores for OOD data. This allows to compute the final scores using the area under the receiver operating characteristic curve (AUC-ROC) to measure the model's ability to detect OOD data. 

\subsection{Model details} 

\textbf{Core architecture.} We use the same feature extractor for both the DUMs architecture. The list of core architectures used across the experiments are: \textit{ResNet18 / ResNet50 / EfficientNet / Swin} \citep{he2016resnet, tan2021effcientnet, liu2021swin} from the torchvision repository \footnote{\url{https://pytorch.org/vision/stable/models.html}} and \textit{Wide-ResNet-28-10} \citep{zagoruyko2016wide} from the original implementation of DUE. Except for the experiment on architecture type and size where ResNet18 has output channels for the residual blocks with size $[64,128,256,512]$,  ResNet18 has output channels for the residual blocks with size $[32, 64, 128, 256]$ which causes small differences in final accuracy.

\textbf{Uncertainty head.} For DUE we use the original implementation \footnote{\url{https://github.com/y0ast/DUE}} with by default we use the RBF kernel function. For NatPN we use the original implementation \footnote{\url{https://github.com/borchero/natural-posterior-network}} but change the uncertainty head with a more expressive density estimator. As seen in Table \ref{tab:normalizing_flow}, we found that a more expressive normalizing flow with resampled base \citep{durkan2019nsf, stimper2022resampled-nf} improves significantly the results over a simpler radial normalizing flow \citep{rezende2015nf} across all the metrics. For all the experiments (except toys where we use radial flow) we use NSF-R with 16 layers.

\begin{table}[!htp]\centering
\caption{\textbf{Normalizing flow expressivity comparison.} Using more expressive normalizing flow significantely improves all the results for NatPN.}
\label{tab:normalizing_flow}
\tiny

\resizebox{0.9\textwidth}{!}{%
\begin{tabular}{lccccccccc}\toprule
&\multicolumn{4}{c}{\textbf{CIFAR100}} &\multicolumn{4}{c}{\textbf{Camelyon}} \\
\cmidrule(lr){2-5}\cmidrule(lr){6-9}
\textbf{Head} &\textbf{Accuracy ($\uparrow$)} &\textbf{Brier Score ($\downarrow$)} &\textbf{OOD Pred. ($\uparrow$)} &\textbf{OOD Epis. ($\uparrow$)} &\textbf{Accuracy ($\uparrow$)} &\textbf{Brier Score ($\downarrow$)} &\textbf{OOD Pred. ($\uparrow$)} &\textbf{OOD Epis. ($\uparrow$)} \\
\midrule
Radial &71.09 $\pm$ 0.21 &52.27 $\pm$ 0.28 &72.84 $\pm$ 1.82 &50.95 $\pm$ 2.16 &83.14 $\pm$ 0.93 &24.55 $\pm$ 1.91 &60.27 $\pm$ 5.29 &69.16 $\pm$ 7.94 \\
NSF-R &\textbf{71.61 $\pm$ 0.07} &\textbf{43.44 $\pm$ 0.11} &\textbf{73.54 $\pm$ 1.69} &\textbf{72.85 $\pm$ 1.25} &\textbf{89.84 $\pm$ 7.93} &\textbf{12.52 $\pm$ 6.17} &\textbf{64.14 $\pm$ 10.42} &\textbf{81.33 $\pm$ 8.78} \\
\bottomrule
\end{tabular}
}
\end{table}

\section{Training for DUMs Details}
\label{subsec:appendix_training}

By default, we first start by only pretraining the core encoder architecture using the cross-entropy loss, before attaching the DUM uncertainty head to the pretrained encoder and continue with the joint training phase. We do not pretrain the encoder for MNIST. we provide further details in \cref{tab:appendix_default_hyper}.
Following the original method in \cite{charpentier2022natpn}, we train the NatPN uncertainty head before (\textit{warmup}) and after (\textit{finetune}) the joint training. In the warmup phase, we use the lambda scheduler increasing linearly from zero to LR head value in \cref{tab:appendix_default_hyper}. In the finetune phase, we use a multistep scheduler that scales the learning rate by 0.2 at 70\% and 90\% of the training starting from the LR head value in \cref{tab:appendix_default_hyper}. We warmup for 0/5/0 and finetune for 60/200/5 epochs for the datasets MNIST/CIFAR/Camelyon respectively.

\begin{table}[!htp]\centering
\caption{\textbf{Default training hyperparameters.} For CIFAR10, CIFAR100 and Camelyon we first pretrain a core encoder architecture for the join training phase. In MNIST we directly joint train given its lower computational cost.}\label{tab:appendix_default_hyper}
\tiny
\resizebox{\textwidth}{!}{%
\begin{tabular}{lllrllllr}
\toprule
\textbf{Dataset} &\textbf{Phase} &\textbf{Encoder} &\textbf{Epochs} &\textbf{\thead{Optimizer \\ Enc. / Head}} &\textbf{\thead{LR \\ Enc. / Head}} &\textbf{\thead{LR scheduler \\ Enc. / Head}} &\textbf{\thead{Weight decay \\ Enc. / Head}} &\textbf{\thead{Latent \\ Dimension}} \\
\midrule
\multirow{3}{*}{MNIST} &Pretrain &- &- &- &- &- &- &- \\
&Joint train (DUE) &ResNet18 &20 &AdamW / AdamW &1e-3 / 1e-4 &cosine $\eta_{min}$=5e-4 / - &1e-6 / 1e-6 &16 \\
&Joint train (NatPN) &ResNet18 &20 &AdamW / AdamW &1e-3 / 5e-3 &cosine $\eta_{min}$=5e-4 / - &1e-6 / 1e-6 &16 \\
\midrule
\multirow{3}{*}{CIFAR10 \& CIFAR100} &Pretrain &ResNet18 &200 &SGD &1e-1 &cosine $\eta_{min}$=5e-4 &5e-4 &- \\
&Joint train (DUE) &ResNet18 &20 &AdamW / AdamW &1e-4 / 1e-4 &cosine $\eta_{min}$=1e-5 / - &1e-6 / 1e-6 &64 \\
&Joint train (NatPN) &ResNet18 &20 &AdamW / AdamW &1e-5 / 5e-3 &cosine $\eta_{min}$=1e-5 / - &1e-6 / 1e-6 &64 \\
\midrule
\multirow{3}{*}{Camelyon} &Pretrain &WideResNet28-10 &5 &AdamW &1e-3 &cosine $\eta_{min}$=1e-5 &1e-8 &- \\
&Joint train (DUE) &WideResNet28-10 &1 &AdamW / AdamW &1e-5 / 5e-3 &cosine $\eta_{min}$=1e-6 / - &1e-6 / 1e-6 &128 \\
&Joint train (NatPN) &WideResNet28-10 &1 &AdamW / AdamW &5e-6 / 1e-5 &cosine $\eta_{min}$=1e-6 / - &1e-6 / 1e-6 &128 \\
\bottomrule
\end{tabular}
}
\end{table}

\textbf{Decoupling learning rate.} In this experiment we use different values for the learning rates of the core architecture and of the uncertainty head. After the decoupling learning rate experiment, we choose the best combination of learning rates through model selection via the validation results and apply it to other experiments. E.g., for the joint training schema and pretraining schema experiments, NatPN uses a learning rate of 1e-4/1e-4 for encoder/head respectively, while for DUE it is 1e-5/5e-3.

\textbf{Training schemes.} In this experiment, we compare the \textit{joint} training in which we jointly train the weights of the core architecture and uncertainty head, and the \textit{sequential} training in which we only train the uncertainty head by keeping the weights of the pretrained core architecture fixed. For each of the setting, we apply two additional techniques to stabilize the training: adding a \textit{batch normalization} to the last layer of the encoder to enforce latent representations to locate in a normalized region \citep{ioffe2015bn,charpentier2022natpn}, and \textit{resetting the last layer} to retrain its weights to improve robustness to spurious correlation \citep{kirichenko2022reset}.

\textbf{Pretraining schemes.} In this experiment, we do not pretrain the core encoder architecture or pretrain it with 10\% of CIFAR100, 100\% of CIFAR100, and ImageNet.
We use ResNet50 as the core architecture with the default setting in \cref{tab:appendix_default_hyper} for CIFAR100 but changing LR to 5e-2. The pretrained model on ImageNet is loaded from the torchvision. The re-scaling transformation applied to CIFAR100 uses the bilinear interpolation, and we add the Gaussian noise with zero mean and variance of 0.1 and 0.5 to the training set to simulate lower quality data.
For the schemes \textit{None} and \textit{C100 (10\%)} which use no or few pretraining data, we increase the joint training phase to 200 epochs with for the core architecture to ensure proper convergence.

\begin{table}[!htb]\centering
\caption{\textbf{Train schema OOD detection.} Uncertainty estimation results broken down for each OOD dataset. We observe that NatPN performs significantly better when using \textit{joint}, while DUE is insensitive to the schema used. The ID dataset used for training is CIFAR100.}
\label{tab:training_schema_ood}
\tiny

\begin{tabular}{lllccc}
\toprule
\textbf{Model} &\textbf{OOD Data} &\textbf{Train Schema} &\textbf{OOD Alea. ($\uparrow$)} &\textbf{OOD Epis. ($\uparrow$)} &\textbf{OOD Pred. ($\uparrow$)} \\
\midrule
\multirow{30}{*}{NatPN} &\multirow{6}{*}{SVHN} &joint &80.64 $\pm$ 1.22 &65.00 $\pm$ 3.18 &80.64 $\pm$ 1.22 \\
& &joint + bn &\textbf{82.07 $\pm$ 0.62} &69.98 $\pm$ 4.40 &\textbf{82.07 $\pm$ 0.62} \\
& &joint + reset &80.77 $\pm$ 1.63 &\textbf{74.67 $\pm$ 2.37} &80.77 $\pm$ 1.63 \\
& &sequential &80.76 $\pm$ 0.95 &43.99 $\pm$ 4.16 &80.76 $\pm$ 0.95 \\
& &sequential + bn &80.64 $\pm$ 0.83 &44.46 $\pm$ 5.63 &80.64 $\pm$ 0.83 \\
& &sequential + reset &78.92 $\pm$ 1.06 &59.54 $\pm$ 4.25 &78.92 $\pm$ 1.06 \\
\cmidrule[0.1pt](lr){2-6}
&\multirow{6}{*}{STL10} &joint &76.63 $\pm$ 0.20 &58.79 $\pm$ 0.24 &76.63 $\pm$ 0.20 \\
& &joint + bn &76.53 $\pm$ 0.22 &63.51 $\pm$ 0.38 &76.53 $\pm$ 0.22 \\
& &joint + reset &76.92 $\pm$ 0.36 &\textbf{64.44 $\pm$ 0.52} &76.92 $\pm$ 0.36 \\
& &sequential &77.57 $\pm$ 0.20 &42.83 $\pm$ 0.62 &77.57 $\pm$ 0.20 \\
& &sequential + bn &\textbf{77.64 $\pm$ 0.22} &44.26 $\pm$ 0.59 &\textbf{77.64 $\pm$ 0.22} \\
& &sequential + reset &77.35 $\pm$ 0.15 &57.56 $\pm$ 0.38 &77.35 $\pm$ 0.15 \\
\cmidrule[0.1pt](lr){2-6}
&\multirow{6}{*}{CelebA} &joint &51.42 $\pm$ 1.29 &28.90 $\pm$ 1.59 &51.42 $\pm$ 1.29 \\
& &joint + bn &51.45 $\pm$ 1.15 &27.67 $\pm$ 2.45 &51.45 $\pm$ 1.15 \\
& &joint + reset &52.01 $\pm$ 0.46 &\textbf{32.54 $\pm$ 0.32} &52.01 $\pm$ 0.46 \\
& &sequential &52.58 $\pm$ 1.08 &24.41 $\pm$ 1.94 &52.58 $\pm$ 1.08 \\
& &sequential + bn &52.44 $\pm$ 1.11 &23.65 $\pm$ 1.85 &52.44 $\pm$ 1.11 \\
& &sequential + reset &\textbf{53.12 $\pm$ 0.92} &26.82 $\pm$ 1.41 &\textbf{53.12 $\pm$ 0.92} \\
\cmidrule[0.1pt](lr){2-6}
&\multirow{6}{*}{Camelyon} &joint &\textbf{67.17 $\pm$ 5.30} &67.03 $\pm$ 9.00 &\textbf{67.17 $\pm$ 5.30} \\
& &joint + bn &61.06 $\pm$ 2.70 &69.69 $\pm$ 5.54 &61.06 $\pm$ 2.70 \\
& &joint + reset &67.07 $\pm$ 1.12 &\textbf{73.45 $\pm$ 4.22} &67.07 $\pm$ 1.12 \\
& &sequential &64.54 $\pm$ 2.08 &56.23 $\pm$ 6.07 &64.54 $\pm$ 2.08 \\
& &sequential + bn &64.34 $\pm$ 2.12 &49.31 $\pm$ 5.99 &64.34 $\pm$ 2.12 \\
& &sequential + reset &63.76 $\pm$ 2.13 &65.59 $\pm$ 4.65 &63.76 $\pm$ 2.13 \\
\cmidrule[0.1pt](lr){2-6}
&\multirow{6}{*}{SVHN OODom.} &joint &100.00 $\pm$ 0.00 &100.00 $\pm$ 0.00 &100.00 $\pm$ 0.00 \\
& &joint + bn &100.00 $\pm$ 0.00 &100.00 $\pm$ 0.00 &100.00 $\pm$ 0.00 \\
& &joint + reset &100.00 $\pm$ 0.00 &100.00 $\pm$ 0.00 &100.00 $\pm$ 0.00 \\
& &sequential &99.99 $\pm$ 0.00 &100.00 $\pm$ 0.00 &99.99 $\pm$ 0.00 \\
& &sequential + bn &100.00 $\pm$ 0.00 &100.00 $\pm$ 0.00 &100.00 $\pm$ 0.00 \\
& &sequential + reset &100.00 $\pm$ 0.00 &100.00 $\pm$ 0.00 &100.00 $\pm$ 0.00 \\

\midrule

\multirow{30}{*}{DUE} &\multirow{6}{*}{SVHN} &joint &- &- &80.75 $\pm$ 0.79 \\
& &joint + bn &- &- &80.47 $\pm$ 1.09 \\
& &joint + reset &- &- &80.92 $\pm$ 0.70 \\
& &sequential &- &- &81.04 $\pm$ 0.82 \\
& &sequential + bn &- &- &\textbf{81.20 $\pm$ 0.86} \\
& &sequential + reset &- &- &80.66 $\pm$ 1.07 \\
\cmidrule[0.1pt](lr){2-6}
&\multirow{6}{*}{STL10} &joint &- &- &77.20 $\pm$ 0.19 \\
& &joint + bn &- &- &77.24 $\pm$ 0.22 \\
& &joint + reset &- &- &77.42 $\pm$ 0.25 \\
& &sequential &- &- &77.50 $\pm$ 0.09 \\
& &sequential + bn &- &- &\textbf{77.52 $\pm$ 0.13} \\
& &sequential + reset &- &- &77.48 $\pm$ 0.17 \\
\cmidrule[0.1pt](lr){2-6}
&\multirow{6}{*}{CelebA} &joint &- &- &47.99 $\pm$ 1.25 \\
& &joint + bn &- &- &47.90 $\pm$ 1.13 \\
& &joint + reset &- &- &49.10 $\pm$ 0.86 \\
& &sequential &- &- &48.07 $\pm$ 0.99 \\
& &sequential + bn &- &- &48.43 $\pm$ 1.06 \\
& &sequential + reset &- &- &\textbf{49.39 $\pm$ 1.29} \\
\cmidrule[0.1pt](lr){2-6}
&\multirow{6}{*}{Camelyon} &joint &- &- &67.76 $\pm$ 2.20 \\
& &joint + bn &- &- &67.54 $\pm$ 2.33 \\
& &joint + reset &- &- &67.03 $\pm$ 2.00 \\
& &sequential &- &- &67.49 $\pm$ 2.58 \\
& &sequential + bn &- &- &67.23 $\pm$ 2.72 \\
& &sequential + reset &- &- &\textbf{67.85 $\pm$ 2.54} \\
\cmidrule[0.1pt](lr){2-6}
&\multirow{6}{*}{SVHN OODom.} &joint &- &- &100.00 $\pm$ 0.00 \\
& &joint + bn &- &- &100.00 $\pm$ 0.00 \\
& &joint + reset &- &- &100.00 $\pm$ 0.00 \\
& &sequential &- &- &100.00 $\pm$ 0.00 \\
& &sequential + bn &- &- &100.00 $\pm$ 0.00 \\
& &sequential + reset &- &- &100.00 $\pm$ 0.00 \\
\bottomrule
\end{tabular}
\end{table}


\begin{table}[!htb]\centering
\caption{\textbf{Pretrain schema OOD detection.} Uncertainty estimation results broken down for each OOD dataset. We see how ImageNet pretrained encoder consistently performs better than other settings for 4/5 OOD datasets. The ID dataset used in the joint training phase is CIFAR100.}
\label{tab:training_pretrain_ood}
\tiny
\begin{tabular}{lllccc}
    \toprule
    \textbf{Model} &\textbf{OOD Data} &\textbf{Pretrain Schema} &\textbf{OOD Alea. ($\uparrow$)} &\textbf{OOD Epis. ($\uparrow$)} &\textbf{OOD Pred. ($\uparrow$)} \\
    \midrule
    \multirow{30}{*}{NatPN} 
    &\multirow{6}{*}{SVHN} &None &79.95 $\pm$ 1.18 &77.87 $\pm$ 2.54 &79.95 $\pm$ 1.18 \\
    & &C100 (10\%) &71.81 $\pm$ 1.67 &75.16 $\pm$ 2.20 &71.81 $\pm$ 1.67 \\
& &C100 (100\%) + $\gN(0.5)$ &80.76 $\pm$ 1.03 &61.15 $\pm$ 6.04 &80.76 $\pm$ 1.03 \\
& &C100 (100\%) + $\gN(0.1)$ &79.86 $\pm$ 1.81 &60.50 $\pm$ 7.54 &79.86 $\pm$ 1.81 \\
    & &C100 (100\%) &81.76 $\pm$ 1.38 &65.72 $\pm$ 4.91 &81.76 $\pm$ 1.38 \\
    & &ImageNet &\textbf{89.34 $\pm$ 0.66} &\textbf{92.02 $\pm$ 0.49} &\textbf{89.34 $\pm$ 0.66} \\
    \cmidrule[0.1pt](lr){2-6}
    &\multirow{6}{*}{STL10} &None &80.34 $\pm$ 0.77 &78.05 $\pm$ 2.37 &80.34 $\pm$ 0.77 \\
    & &C100 (10\%) &74.64 $\pm$ 0.72 &62.43 $\pm$ 2.92 &74.64 $\pm$ 0.72 \\
& &C100 (100\%) + $\gN(0.5)$ &79.92 $\pm$ 0.32 &56.81 $\pm$ 3.69 &79.92 $\pm$ 0.32 \\
& &C100 (100\%) + $\gN(0.1)$ &80.63 $\pm$ 0.94 &57.72 $\pm$ 3.93 &80.63 $\pm$ 0.94 \\
    & &C100 (100\%) &80.70 $\pm$ 1.33 &66.07 $\pm$ 2.60 &80.70 $\pm$ 1.33 \\
    & &ImageNet &\textbf{85.46 $\pm$ 0.50} &\textbf{90.76 $\pm$ 0.31} &\textbf{85.46 $\pm$ 0.50} \\
    \cmidrule[0.1pt](lr){2-6}
    &\multirow{6}{*}{CelebA} &None &73.59 $\pm$ 3.78 &\textbf{76.19 $\pm$ 3.32} &73.59 $\pm$ 3.78 \\
    & &C100 (10\%) &73.26 $\pm$ 1.79 &63.31 $\pm$ 3.66 &73.26 $\pm$ 1.79 \\
& &C100 (100\%) + $\gN(0.5)$ &66.88 $\pm$ 2.72 &49.00 $\pm$ 3.23 &66.88 $\pm$ 2.72 \\
& &C100 (100\%) + $\gN(0.1)$ &73.15 $\pm$ 0.92 &46.91 $\pm$ 6.50 &73.15 $\pm$ 0.92 \\
    & &C100 (100\%) &\textbf{73.61 $\pm$ 2.16} &58.11 $\pm$ 2.85 &\textbf{73.61 $\pm$ 2.16} \\
    & &ImageNet &59.30 $\pm$ 3.77 &64.80 $\pm$ 2.15 &59.30 $\pm$ 3.77 \\
    \cmidrule[0.1pt](lr){2-6}
    &\multirow{6}{*}{Camelyon} &None &65.37 $\pm$ 10.77 &96.95 $\pm$ 2.01 &65.37 $\pm$ 10.77 \\
    & &C100 (10\%) &20.84 $\pm$ 5.27 &91.58 $\pm$ 7.56 &20.84 $\pm$ 5.27 \\
& &C100 (100\%) + $\gN(0.5)$ &40.19 $\pm$ 5.07 &76.33 $\pm$ 5.84 &40.19 $\pm$ 5.07 \\
& &C100 (100\%) + $\gN(0.1)$ &45.39 $\pm$ 10.58 &83.79 $\pm$ 5.27 &45.39 $\pm$ 10.58 \\
    & &C100 (100\%) &58.69 $\pm$ 11.10 &94.63 $\pm$ 2.85 &58.69 $\pm$ 11.10 \\
    & &ImageNet &\textbf{90.64 $\pm$ 2.47} &\textbf{97.82 $\pm$ 0.56} &\textbf{90.64 $\pm$ 2.47} \\
    \cmidrule[0.1pt](lr){2-6}
    &\multirow{6}{*}{SVHN OODom.} &None &100.00 $\pm$ 0.00 &100.00 $\pm$ 0.00 &100.00 $\pm$ 0.00 \\
    & &C100 (10\%) &99.96 $\pm$ 0.04 &100.00 $\pm$ 0.00 &99.96 $\pm$ 0.04 \\
& &C100 (100\%) + $\gN(0.5)$ &99.94 $\pm$ 0.06 &100.00 $\pm$ 0.00 &99.94 $\pm$ 0.06 \\
& &C100 (100\%) + $\gN(0.1)$ &100.00 $\pm$ 0.00 &100.00 $\pm$ 0.00 &100.00 $\pm$ 0.00 \\
    & &C100 (100\%) &100.00 $\pm$ 0.00 &100.00 $\pm$ 0.00 &100.00 $\pm$ 0.00 \\
    & &ImageNet &\textbf{100.00 $\pm$ 0.00} &\textbf{100.00 $\pm$ 0.00} &\textbf{100.00 $\pm$ 0.00} \\
    \midrule
    \multirow{30}{*}{DUE} &\multirow{6}{*}{SVHN} &None &- &- &82.66 $\pm$ 0.77 \\
    & &C100 (10\%) &- &- &77.24 $\pm$ 0.53 \\
& &C100 (100\%) + $\gN(0.5)$ &- &- &80.61 $\pm$ 1.01 \\
& &C100 (100\%) + $\gN(0.1)$ &- &- &77.94 $\pm$ 1.18 \\
    & &C100 (100\%) &- &- &78.76 $\pm$ 1.74 \\
    & &ImageNet &- &- &\textbf{92.18 $\pm$ 0.11} \\
    \cmidrule[0.1pt](lr){2-6}
    &\multirow{6}{*}{STL10} &None &- &- &78.91 $\pm$ 1.06 \\
    & &C100 (10\%) &- &- &75.98 $\pm$ 0.87 \\
& &C100 (100\%) + $\gN(0.5)$ &- &- &79.49 $\pm$ 0.33 \\
& &C100 (100\%) + $\gN(0.1)$ &- &- &80.21 $\pm$ 0.83 \\
    & &C100 (100\%) &- &- &79.63 $\pm$ 1.34 \\
    & &ImageNet &- &- &\textbf{89.56 $\pm$ 0.53} \\
    \cmidrule[0.1pt](lr){2-6}
    &\multirow{6}{*}{CelebA} &None &- &- &65.64 $\pm$ 1.77 \\
    & &C100 (10\%) &- &- &\textbf{74.62 $\pm$ 1.56} \\
& &C100 (100\%) + $\gN(0.5)$ &- &- &66.58 $\pm$ 3.20 \\
& &C100 (100\%) + $\gN(0.1)$ &- &- &75.15 $\pm$ 1.84 \\
    & &C100 (100\%) &- &- &74.60 $\pm$ 1.11 \\
    & &ImageNet &- &- &72.19 $\pm$ 1.42 \\
    \cmidrule[0.1pt](lr){2-6}
    &\multirow{6}{*}{Camelyon} &None &- &- &73.00 $\pm$ 2.81 \\
    & &C100 (10\%) &- &- &34.84 $\pm$ 3.53 \\
& &C100 (100\%) + $\gN(0.5)$ &- &- &47.78 $\pm$ 5.30 \\
& &C100 (100\%) + $\gN(0.1)$ &- &- &63.88 $\pm$ 5.79 \\
    & &C100 (100\%) &- &- &75.58 $\pm$ 5.17 \\
    & &ImageNet &- &- &\textbf{97.28 $\pm$ 0.47} \\
    \cmidrule[0.1pt](lr){2-6}
    &\multirow{6}{*}{SVHN OODom.} &None &- &- &100.00 $\pm$ 0.00 \\
    & &C100 (10\%) &- &- &99.51 $\pm$ 0.12 \\
& &C100 (100\%) + $\gN(0.5)$ &- &- &99.99 $\pm$ 0.00 \\
& &C100 (100\%) + $\gN(0.1)$ &- &- &99.99 $\pm$ 0.00 \\
    & &C100 (100\%) &- &- &99.99 $\pm$ 0.00 \\
    & &ImageNet &- &- &\textbf{100.00 $\pm$ 0.00} \\
    \bottomrule
\end{tabular}
\end{table}

\section{Architecture for DUMs Details}
\label{subsec:appendix_encoder}



For all the architecture experiments we used the default training settings in \cref{subsec:appendix_training} and \cref{tab:appendix_default_hyper}.

\textbf{Bi-Lipschitz training details.} In the \textit{None} configuration, we removed the residual connection from the architecture for both the pretraining of the encoder and the joint training phase. In the \textit{Residual} configuration, we did not modify anything since both ResNet and WideResNet already use the residual connection. While for the \textit{bi-Lipschitz} configuration, we added the spectral normalization during both the pretraining and also joint training phase. Following the original method presented in \cite{van2021due}, we used the same implementation and applied spectral normalization to the linear, convolution, and batch normalization layers. During the model selection, using the validation set results, we find that the best \textit{Lipschitz constant} $c$ for DUE is 4, and for NatPN is 5. For both the models the power iteration parameter is set to 1. For the toy dataset we use an encoder with 4 linear layers of 128 dimension each.

\begin{figure}[!htb]
    \centering
    \includegraphics[width=\textwidth]{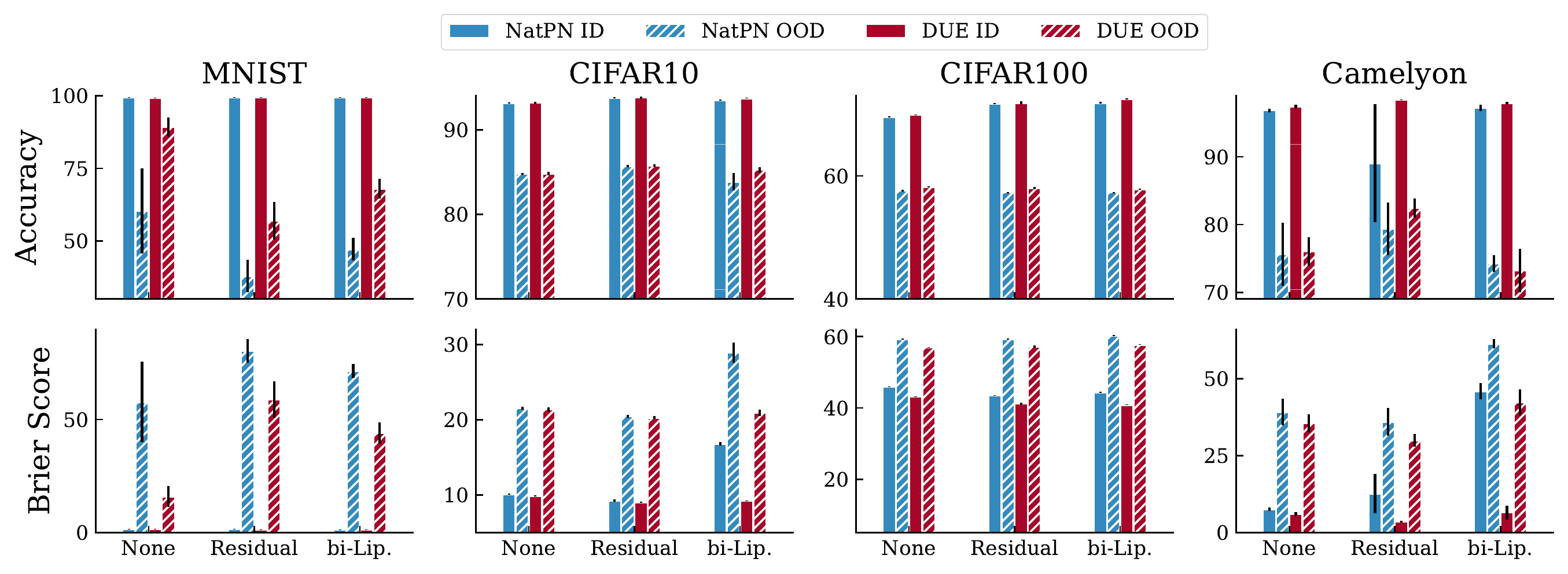}
    \caption{Results OOD generalization and OOD detection results of DUMs with none, residual and bi-lipschitz \textbf{architecture constraints}. Bi-lipschitz and more specifically can improve OOD detection by mitigating feature collapse (see \cref{fig:bi_toy}) at the expense of degrading OOD generalization.}
    \label{fig:bi_bar_full}
\end{figure}

\begin{figure}[!htb]
    \centering
    \includegraphics[width=0.75\textwidth]{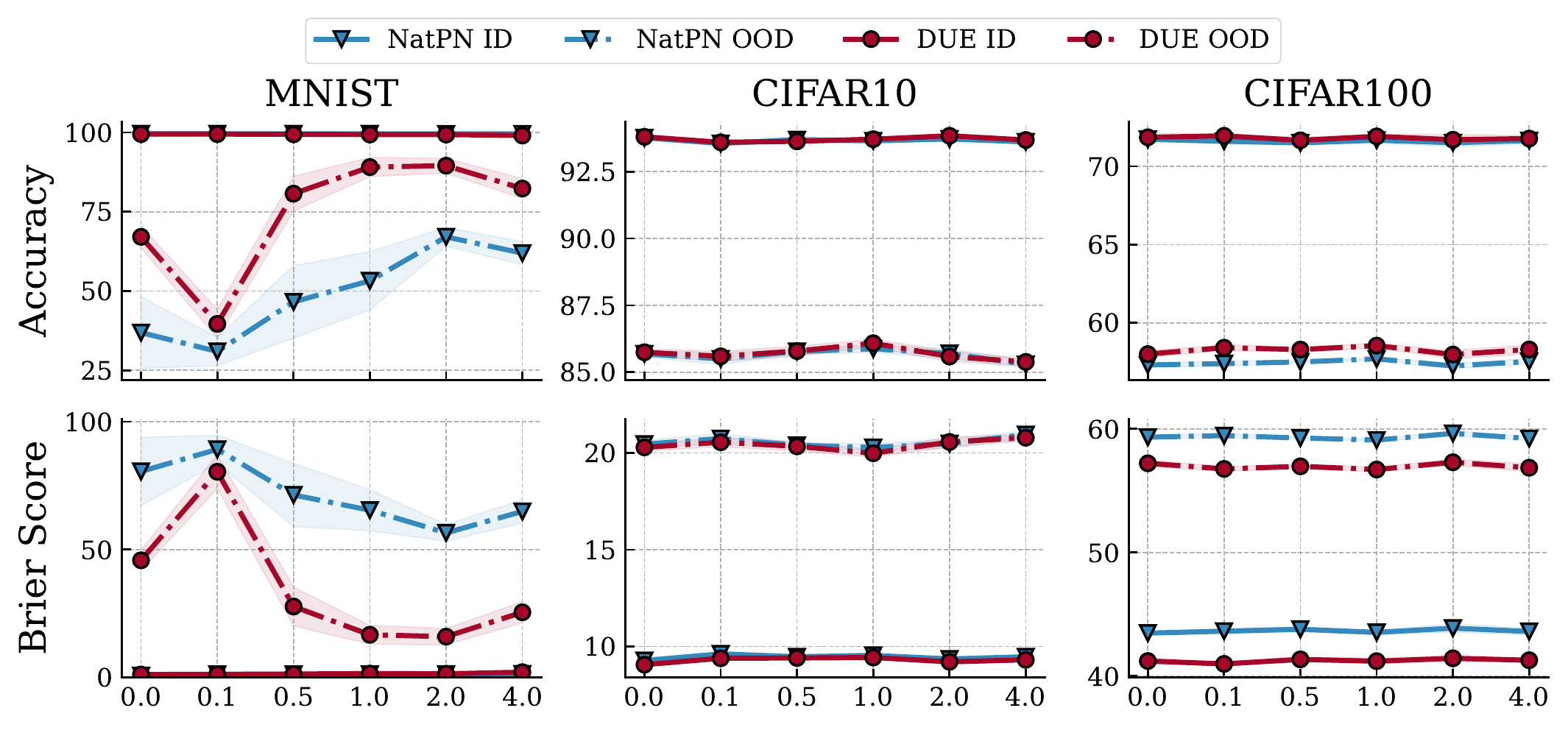}
    \caption{Results OOD generalization and OOD detection results of DUMs with reconstruction \textbf{architecture constraints}. Increasing the strength of the reconstruction factor $\lambda$ improves the OOD generalization only on the simpler MNIST/CMNIST datasets but fails for more complex datasets. }
    \label{fig:reconst_full}
\end{figure}

\textbf{Reconstruction training details.} The decoder reconstructs the input extracted from the last residual block of the encoder, before the pooling layer. During the pretraining phase, both the encoder and decoder are trained with the cross-entropy loss plus a MSE reconstruction term. During the joint training phase, we load the pretrained encoder and decoder, and joint train with the DUMs' respective loss plus the MSE reconstruction term. For the toy dataset we use an encoder with 4 linear layers of 128 dimension each.

\begin{figure}[!htb]
    \begin{subfigure}[b]{\textwidth}
        \centering
        \includegraphics[width=0.75\textwidth]{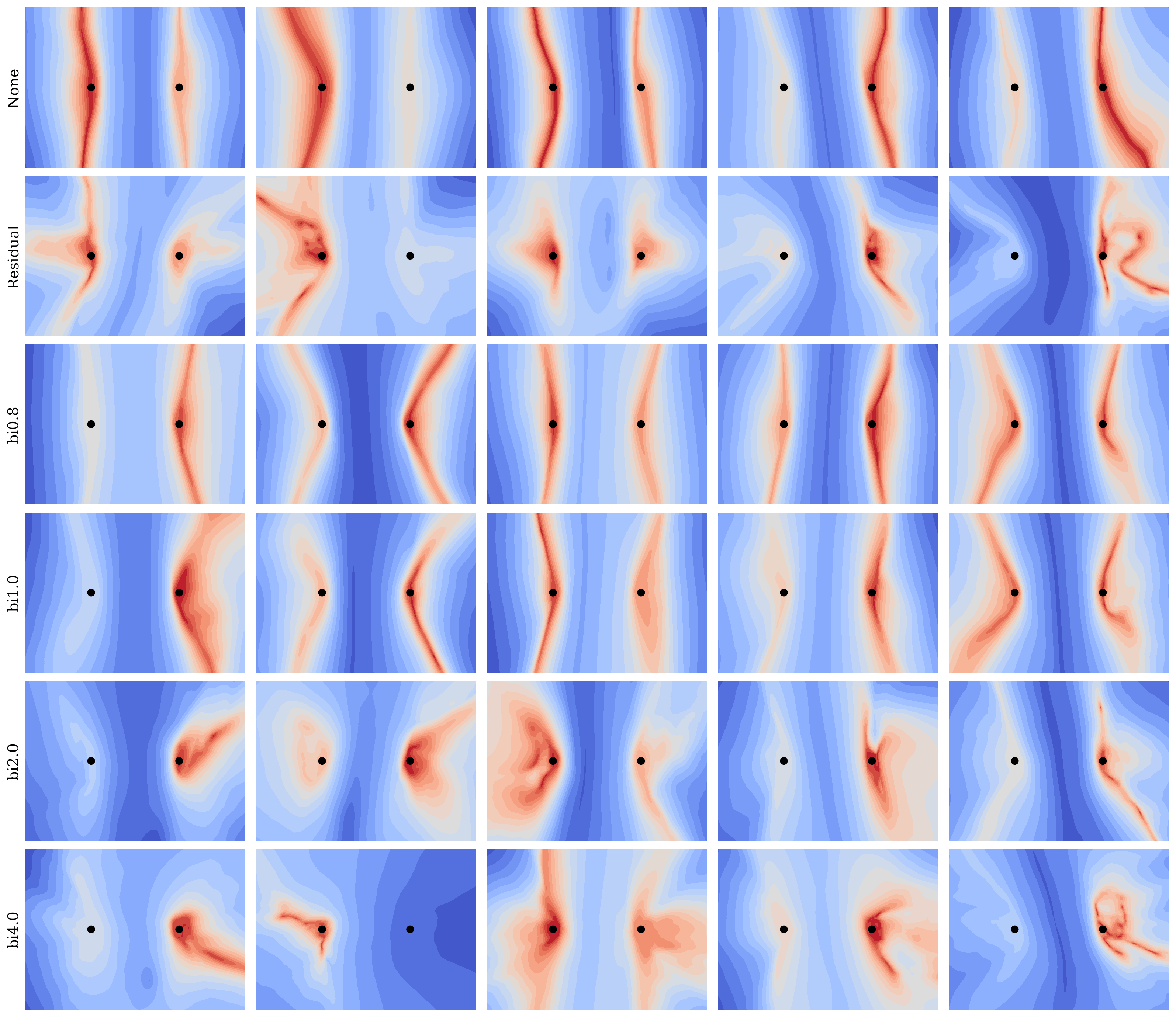}
        \caption{Bi-Lipschitz}
        \label{fig:bi_toy}
    \end{subfigure}
    \begin{subfigure}[b]{\textwidth}
        \centering
        \includegraphics[width=0.75\textwidth]{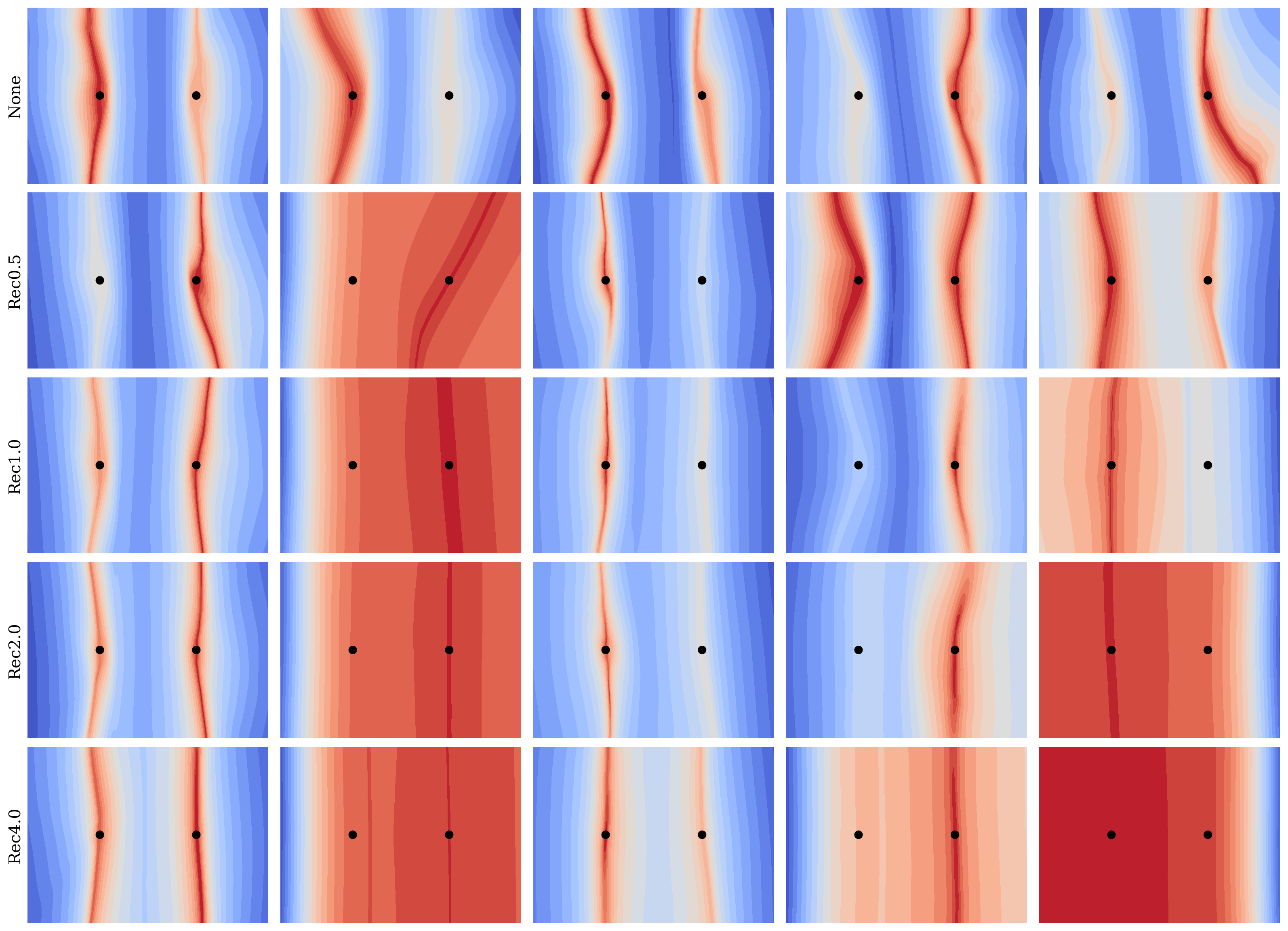}
        \caption{Reconstruction}
        \label{fig:rec_toy}
    \end{subfigure}
    \caption{\textbf{Regularization constraint toy dataset uncertainty boundaries with NatPN.} The two black dots represent the center of two different class of Gaussian data, sharing the same y-axis to trigger the \textit{feature collapse} phenomenon. The color represents the likelihood produced by the uncertainty head. Each row is a different setting, e.g. \textit{bi1.0} is the bi-Lipschitz constraint with the Lipschitz constant $c = 1$ and \textit{rec1.0} is the reconstruction term with $\lambda=1$. Each column is a different seed initialization. (top) \textbf{Bi-Lipschitz} experiment shows that the core encoder architecture constrained with a larger \textit{Lipschitz constant} in the last two rows behaves similar to the encoder constrained with only the residual connection (second row) showing that relaxing the spectral normalization constraint falls back to the residual connection, preventing the feature collapse. (bottom) \textbf{Reconstruction} experiment shows that it does not help to prevent feature collapse by itself. The core encoder architecture is not constrained with bi-Lipschitz. }
    \label{fig:toy}
\end{figure}

\begin{figure}[!htb]
    \begin{subfigure}[b]{\textwidth}
        \centering
        \includegraphics[width=0.8\textwidth]{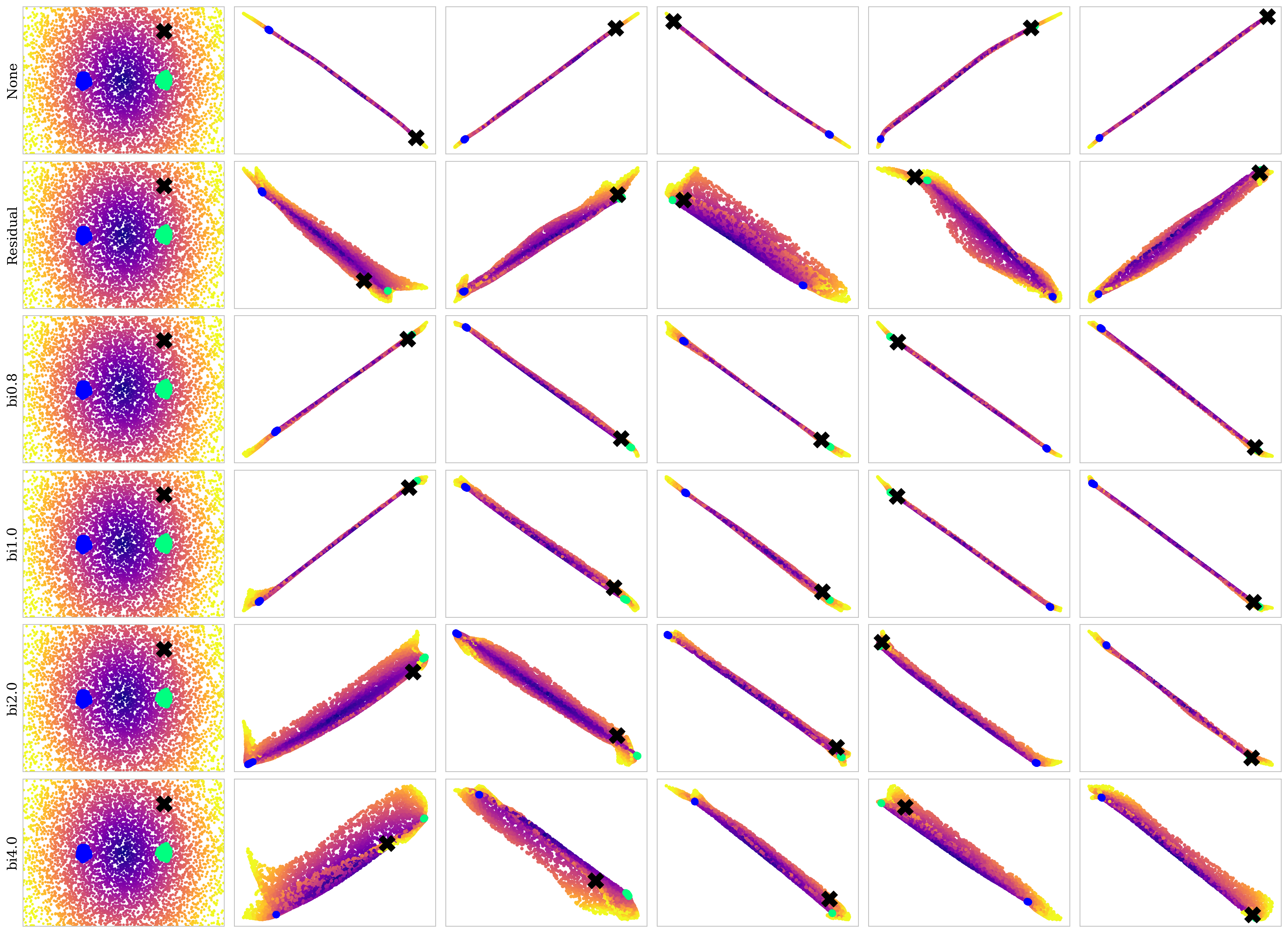}
        \caption{Bi-Lipschitz}
        \label{fig:bi_toy_collapse}
    \end{subfigure}
    \begin{subfigure}[b]{\textwidth}
        \centering
        \includegraphics[width=0.8\textwidth]{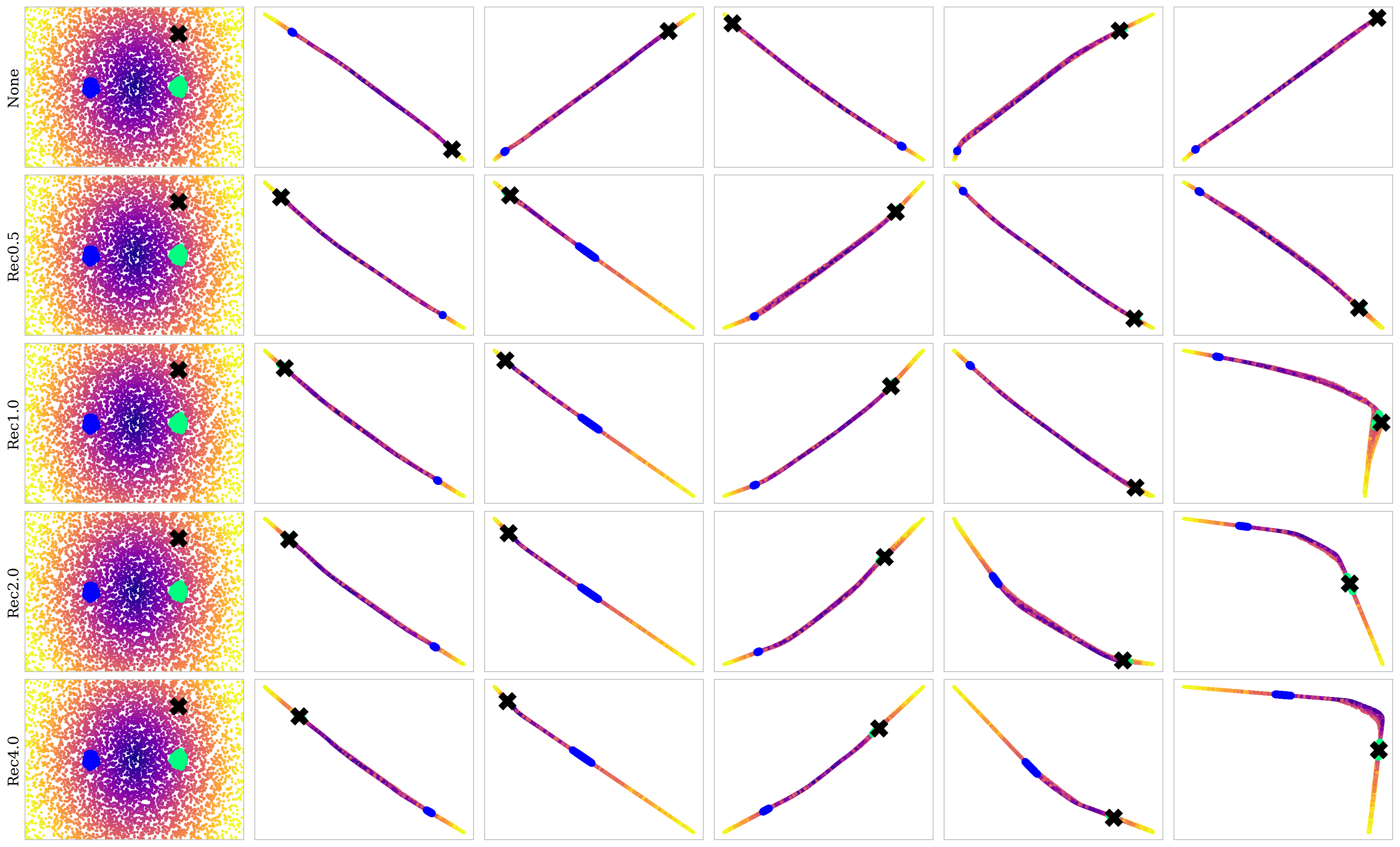}
        \caption{Reconstruction}
        \label{fig:rec_toy_collapse}
    \end{subfigure}
    \caption{\textbf{Regularization constraint toy dataset feature collapse with NatPN.} Similarly to \citep{van2021due}, we run the toy experiment where the first column represent two Gaussian class of data, sharing the same y-axis center to trigger the \textit{feature collapse} phenomenon, and a grid of unrelated point to simulate the space distorsion (colors are based on the Gaussian's generating distribution). 
    Each row is a different setting, e.g. \textit{bi1.0} is the bi-Lipschitz constraint with the Lipschitz constant $c = 1$ and \textit{rec1.0} is the reconstruction term with $\lambda=1$. Each column is a different seed initialization. Results are the same as \ref{fig:bi_toy}. Larger Lipschitz constant \textit{c} reverts back to the residual connection, and reconstruction regularization collapses the 2D dimension into one single dimension. }
    \label{fig:toy_collpase}
\end{figure}


\begin{figure}[!htb]
    \centering
    \begin{subfigure}[b]{\textwidth}
        \includegraphics[width=\textwidth]{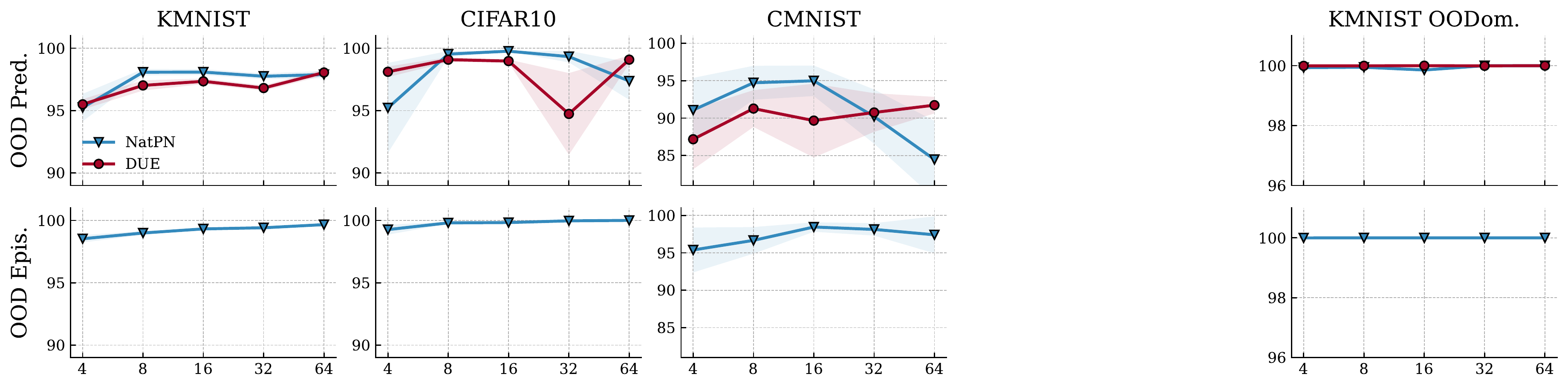}
        \caption{MNIST}
        \label{fig:latent_ood_mnist}
    \end{subfigure}
    \begin{subfigure}[b]{\textwidth}
        \includegraphics[width=\textwidth]{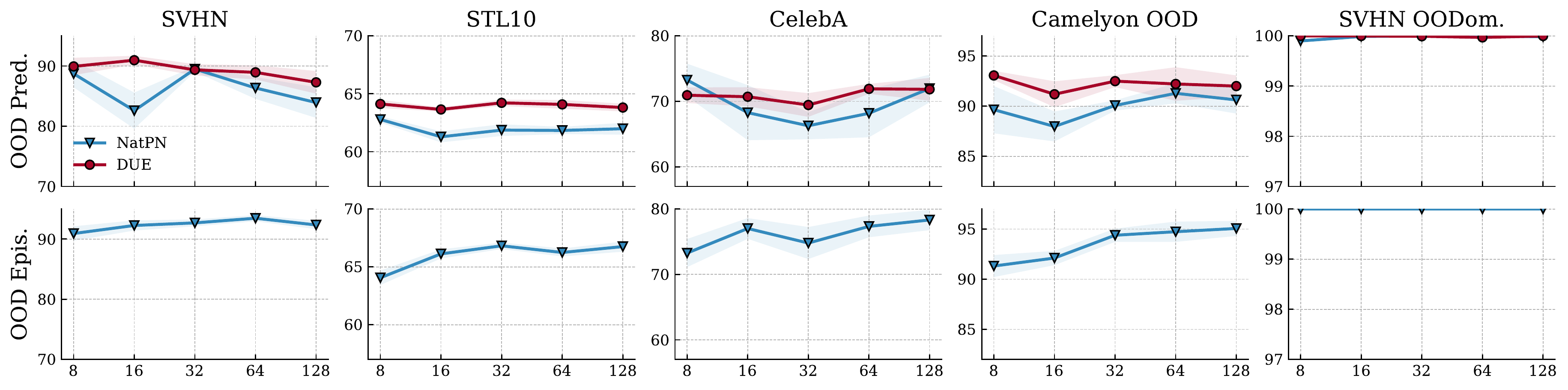}
        \caption{CIFAR10}
        \label{fig:latent_ood_cifar10}
    \end{subfigure}
    \begin{subfigure}[b]{\textwidth}
        \includegraphics[width=\textwidth]{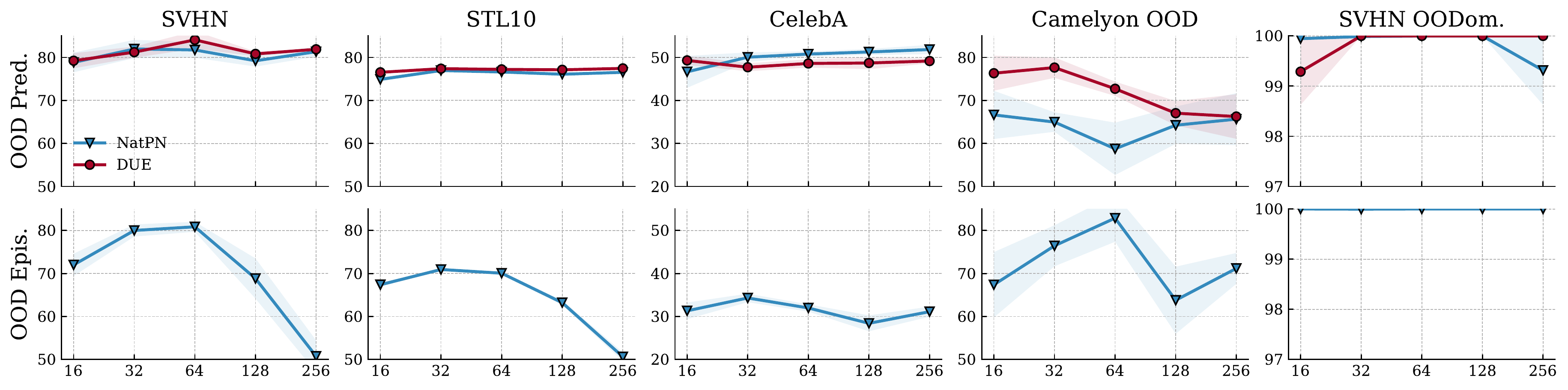}
        \caption{CIFAR100}
        \label{fig:latent_ood_cifar100}
    \end{subfigure}
    \begin{subfigure}[b]{\textwidth}
        \includegraphics[width=\textwidth]{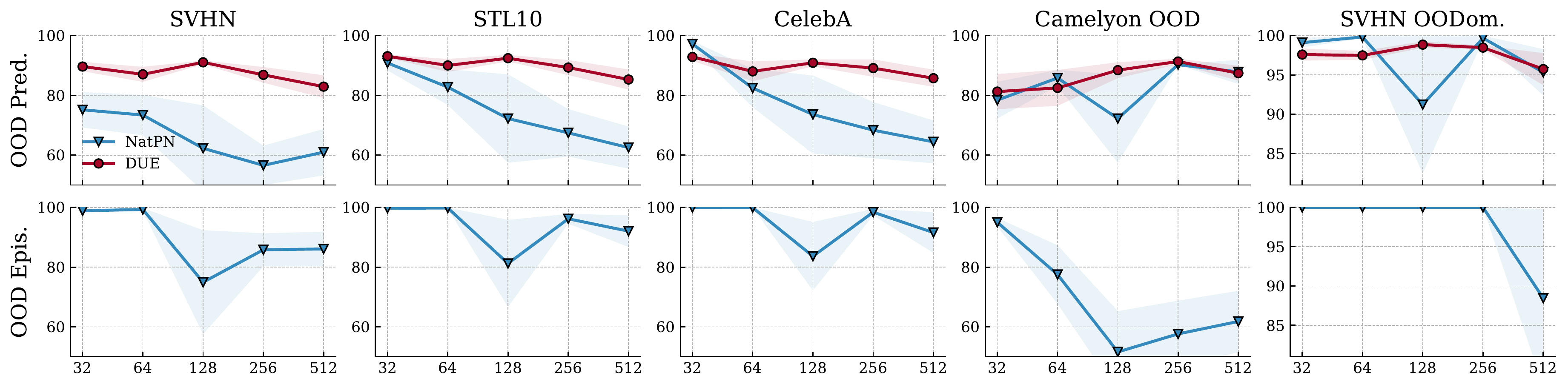}
        \caption{Camelyon ID}
        \label{fig:latent_ood_camelyon}
    \end{subfigure}
    
    \caption{\textbf{Latent dimension OOD detection.} For each training dataset we show the uncertainty estimation results on the corresponding OOD dataset. NatPN encounters numerical instabilities with high latent dimension on Camelyon dataset, while DUE is less sensitive to the variation.}
    \label{fig:latent_ood}
\end{figure}

\begin{figure}[!htb]
    \centering
    \begin{subfigure}[b]{\textwidth}
        \includegraphics[width=\textwidth]{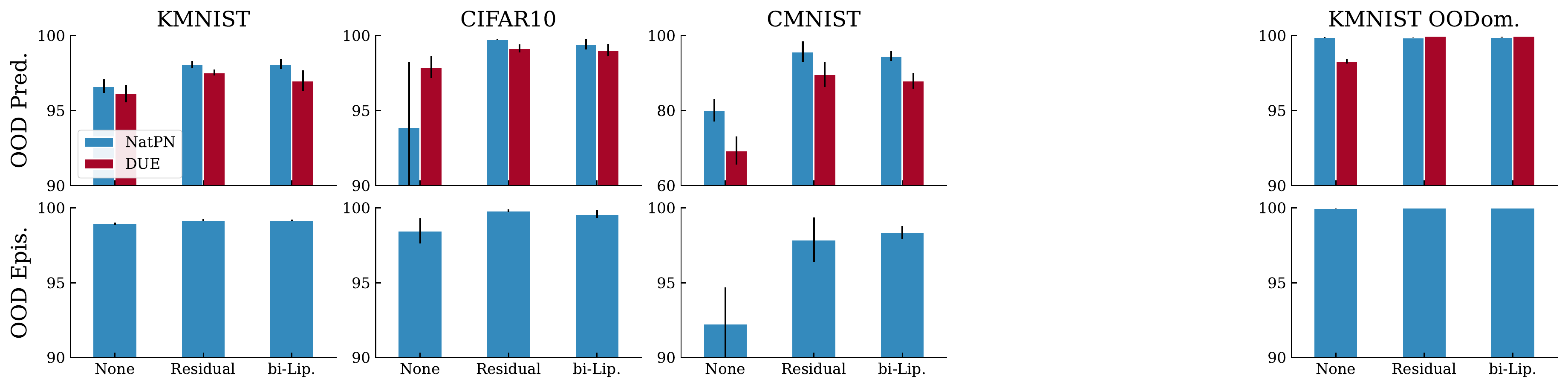}
        \caption{MNIST}
        \label{fig:bi_bar_ood_mnist}
    \end{subfigure}
    \begin{subfigure}[b]{\textwidth}
        \includegraphics[width=\textwidth]{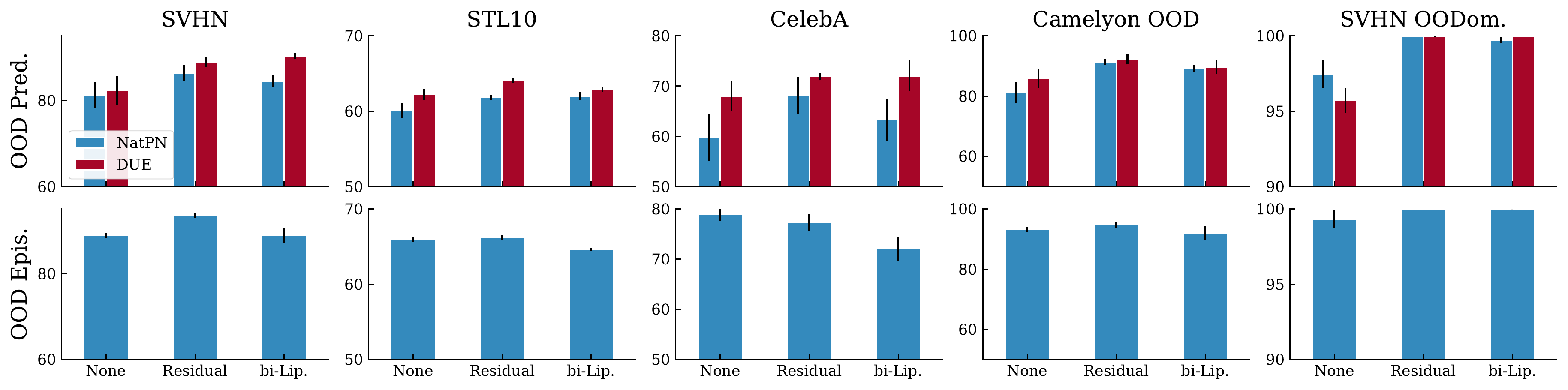}
        \caption{CIFAR10}
        \label{fig:bi_bar_ood_cifar10}
    \end{subfigure}
    \begin{subfigure}[b]{\textwidth}
        \includegraphics[width=\textwidth]{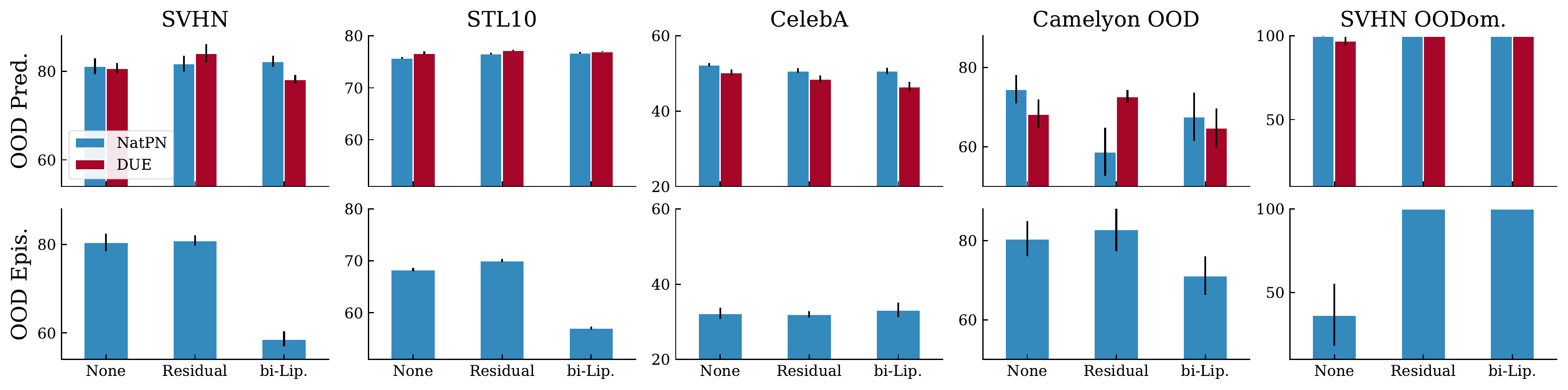}
        \caption{CIFAR100}
        \label{fig:bi_bar_ood_cifar100}
    \end{subfigure}
    \begin{subfigure}[b]{\textwidth}
        \includegraphics[width=\textwidth]{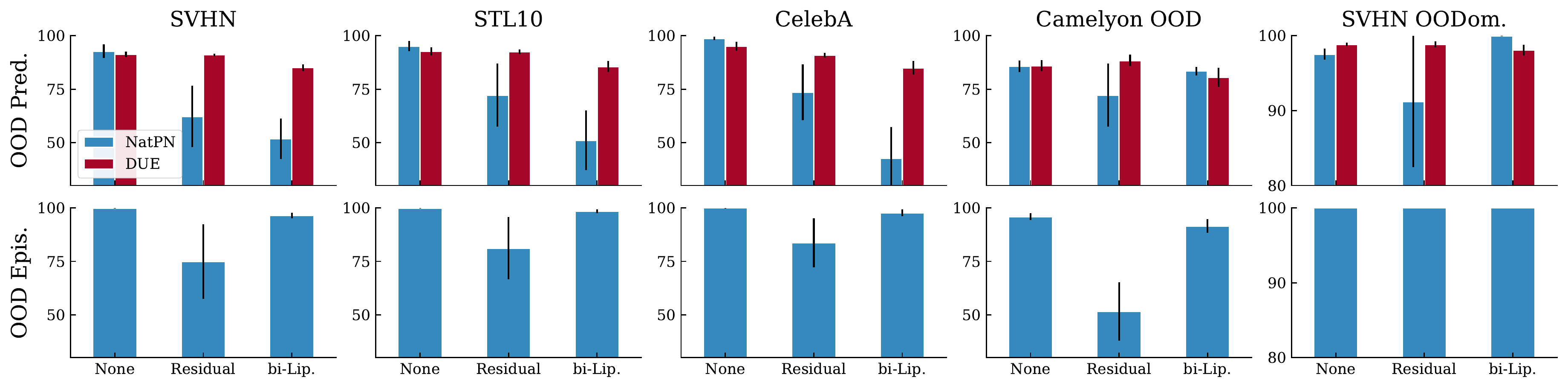}
        \caption{Camelyon ID}
        \label{fig:bi_bar_ood_camelyon}
    \end{subfigure}
    
    \caption{\textbf{Bi-lipschitz OOD detection.} For each training dataset we show the uncertainty estimation results on the corresponding OOD dataset. Bi-Lipschitz improvements are not consistent across different OOD datasets. }
    \label{fig:bi_ood}
\end{figure}

\begin{figure}[!htb]
    \centering
    \begin{subfigure}[b]{\textwidth}
        \includegraphics[width=\textwidth]{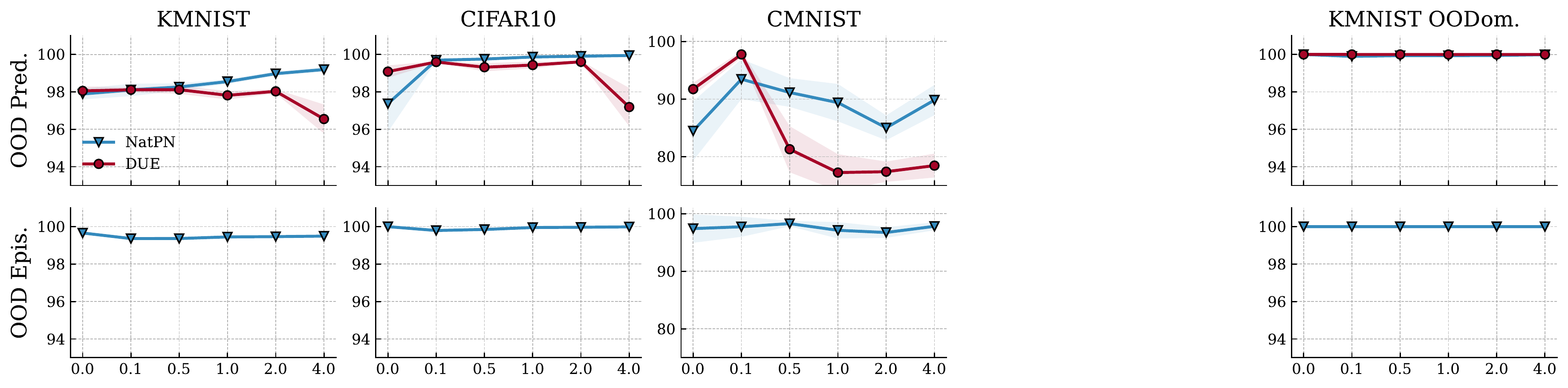}
        \caption{MNIST}
        \label{fig:rec_ood_mnist}
    \end{subfigure}
    \begin{subfigure}[b]{\textwidth}
        \includegraphics[width=\textwidth]{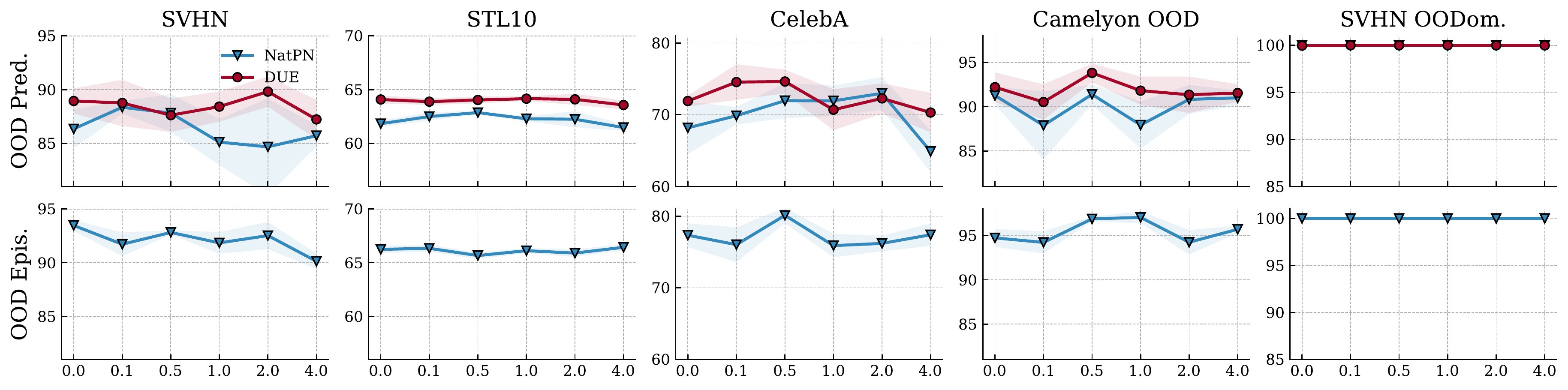}
        \caption{CIFAR10}
        \label{fig:rec_ood_cifar10}
    \end{subfigure}
    \begin{subfigure}[b]{\textwidth}
        \includegraphics[width=\textwidth]{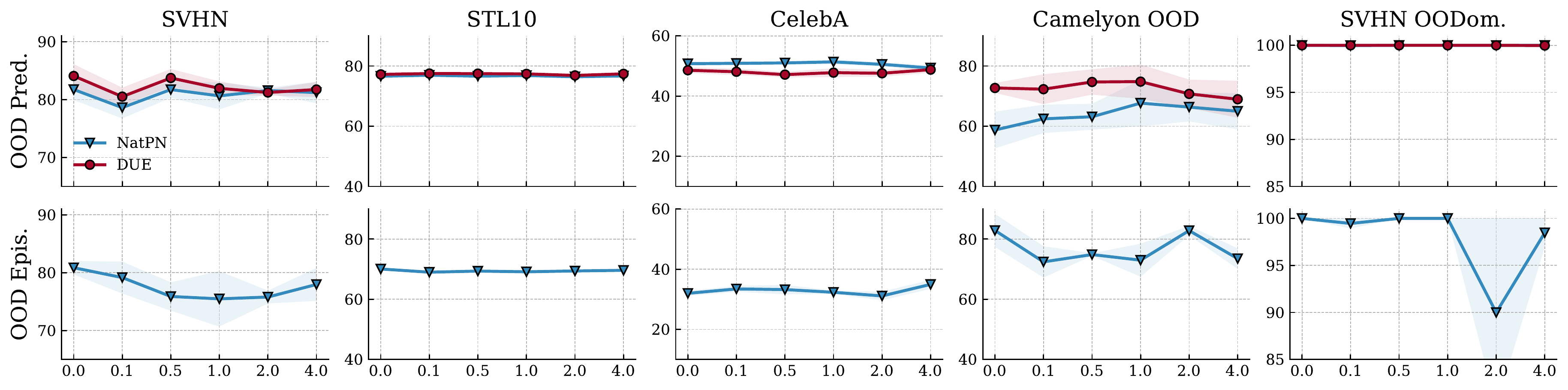}
        \caption{CIFAR100}
        \label{fig:rec_ood_cifar100}
    \end{subfigure}
    
    \caption{\textbf{Reconstruction regularization OOD detection.} Increasing the weight coefficient of the reconstruction loss term improved the OOD detection of NatPN in MNIST. However, we did not observe improvements on more complex datasets such as CIFAR.}
    \label{fig:rec_ood}
\end{figure}

\begin{figure}[!htb]
    \centering
    \includegraphics[width=0.8\linewidth]{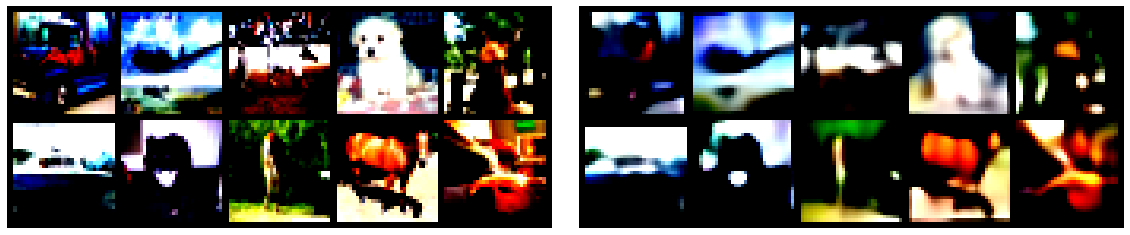}
    \caption{\textbf{Reconstruction regularization CIFAR samples.} (left) The original input (right) the reconstructed input after NatPN's joint training phase. The reconstruction discards detailed information compared to the original input. }
    \label{fig:rec_samples}
\end{figure}

\begin{table}[!htb]
\centering
\caption{\textbf{Encoder architecture OOD detection.} For each training dataset we show the uncertainty estimation results on the corresponding OOD dataset. We observe that new architectures (EfficientNet, Swin) have consistently better results. Interestingly, the transformer based model Swin, is not able to detect out-of-domain data, which should be easy in principle.}
\label{tab:encoder_architecture_ood}
\tiny
\begin{tabular}{lllccc}
    \toprule
    \textbf{Model} &\textbf{OOD Data} &\textbf{Architecture} &\textbf{OOD Alea. ($\uparrow$)} &\textbf{OOD Epis. ($\uparrow$)} &\textbf{OOD Pred. ($\uparrow$)} \\
    \midrule
    \multirow{20}{*}{NatPN} &\multirow{4}{*}{SVHN} &ResNet18 &\textbf{89.90 $\pm$ 1.22} &78.14 $\pm$ 3.13 &\textbf{89.90 $\pm$ 1.22} \\
    & &ResNet50 &89.34 $\pm$ 0.66 &92.02 $\pm$ 0.49 &89.34 $\pm$ 0.66 \\
    & &EfficientNet\_V2\_S &88.65 $\pm$ 0.68 &92.52 $\pm$ 0.60 &88.65 $\pm$ 0.68 \\
    & &Swin\_T &87.73 $\pm$ 1.36 &\textbf{94.17 $\pm$ 0.74} &87.73 $\pm$ 1.36 \\
    \cmidrule[0.1pt](lr){2-6}
    &\multirow{4}{*}{STL10} &ResNet18 &\textbf{90.99 $\pm$ 0.37} &83.90 $\pm$ 0.59 &\textbf{90.99 $\pm$ 0.37} \\
    & &ResNet50 &85.46 $\pm$ 0.50 &90.76 $\pm$ 0.31 &85.46 $\pm$ 0.50 \\
    & &EfficientNet\_V2\_S &88.68 $\pm$ 0.62 &91.11 $\pm$ 0.47 &88.68 $\pm$ 0.62 \\
    & &Swin\_T &85.44 $\pm$ 0.56 &\textbf{92.16 $\pm$ 0.40} &85.44 $\pm$ 0.56 \\
    \cmidrule[0.1pt](lr){2-6}
    &\multirow{4}{*}{CelebA} &ResNet18 &66.46 $\pm$ 3.48 &50.61 $\pm$ 2.14 &66.46 $\pm$ 3.48 \\
    & &ResNet50 &59.30 $\pm$ 3.77 &64.80 $\pm$ 2.15 &59.30 $\pm$ 3.77 \\
    & &EffNet\_V2\_S &\textbf{67.04 $\pm$ 1.74} &66.29 $\pm$ 1.45 &\textbf{67.04 $\pm$ 1.74} \\
    & &Swin\_T &63.60 $\pm$ 2.16 &\textbf{72.80 $\pm$ 1.00} &63.60 $\pm$ 2.16 \\
    \cmidrule[0.1pt](lr){2-6}
    &\multirow{4}{*}{Camelyon} &ResNet18 &90.09 $\pm$ 4.43 &96.28 $\pm$ 1.06 &90.09 $\pm$ 4.43 \\
    & &ResNet50 &90.64 $\pm$ 2.47 &97.82 $\pm$ 0.56 &90.64 $\pm$ 2.47 \\
    & &EfficientNet\_V2\_S &94.56 $\pm$ 0.79 &97.42 $\pm$ 0.44 &94.56 $\pm$ 0.79 \\
    & &Swin\_T &\textbf{95.43 $\pm$ 1.47} &\textbf{98.71 $\pm$ 0.50} &\textbf{95.43 $\pm$ 1.47} \\
    \cmidrule[0.1pt](lr){2-6}
    &\multirow{4}{*}{SVHN OODom.} &ResNet18 &\textbf{100.00 $\pm$ 0.00} &\textbf{100.00 $\pm$ 0.00} &\textbf{100.00 $\pm$ 0.00} \\
    & &ResNet50 &\textbf{100.00 $\pm$ 0.00} &\textbf{100.00 $\pm$ 0.00} &\textbf{100.00 $\pm$ 0.00} \\
    & &EfficientNet\_V2\_S &\textbf{100.00 $\pm$ 0.00} &\textbf{100.00 $\pm$ 0.00} &\textbf{100.00 $\pm$ 0.00} \\
    & &Swin\_T &97.37 $\pm$ 0.29 &93.31 $\pm$ 1.42 &97.37 $\pm$ 0.29 \\
    \midrule
    \multirow{20}{*}{DUE} &\multirow{4}{*}{SVHN} &ResNet18 &- &- &88.77 $\pm$ 0.28 \\
    & &ResNet50 &- &- &92.18 $\pm$ 0.11 \\
    & &EfficientNet\_V2\_S &- &- &90.95 $\pm$ 0.53 \\
    & &Swin\_T &- &- &\textbf{93.62 $\pm$ 0.39} \\
    \cmidrule[0.1pt](lr){2-6}
    &\multirow{4}{*}{STL10} &ResNet18 &- &- &\textbf{90.66 $\pm$ 0.48} \\
    & &ResNet50 &- &- &89.56 $\pm$ 0.53 \\
    & &EfficientNet\_V2\_S &- &- &89.08 $\pm$ 0.39 \\
    & &Swin\_T &- &- &89.73 $\pm$ 0.36 \\
    \cmidrule[0.1pt](lr){2-6}
    &\multirow{4}{*}{CelebA} &ResNet18 &- &- &64.75 $\pm$ 1.44 \\
    & &ResNet50 &- &- &72.19 $\pm$ 1.42 \\
    & &EffNet\_V2\_S &- &- &\textbf{72.28 $\pm$ 1.76} \\
    & &Swin\_T &- &- &69.37 $\pm$ 0.67 \\
    \cmidrule[0.1pt](lr){2-6}
    &\multirow{4}{*}{Camelyon} &ResNet18 &- &- &96.01 $\pm$ 1.13 \\
    & &ResNet50 &- &- &97.28 $\pm$ 0.47 \\
    & &EfficientNet\_V2\_S &- &- &94.82 $\pm$ 0.65 \\
    & &Swin\_T &- &- &\textbf{99.33 $\pm$ 0.14} \\
    \cmidrule[0.1pt](lr){2-6}
    &\multirow{4}{*}{SVHN OODom.} &ResNet18 &- &- &\textbf{100.00 $\pm$ 0.00} \\
    & &ResNet50 &- &- &\textbf{100.00 $\pm$ 0.00} \\
    & &EfficientNet\_V2\_S &- &- &\textbf{100.00 $\pm$ 0.00} \\
    & &Swin\_T &- &- &97.47 $\pm$ 0.22 \\
    \bottomrule
\end{tabular}
\end{table}

\section{Prior for DUMs Details}
\label{subsec:appendix_prior}

For all the prior experiments we use the default training settings in \cref{subsec:appendix_training} and \cref{tab:appendix_default_hyper}. In the following experiments, we vary the \textit{entropy regularization} $\lambda$ and the \textit{evidence prior} $n^{prior}$ for NatPN, and the choice of kernel for DUE. 

Before starting the training, we inject the artificial aleatoric noise by reassigning the target $y$ with a randomly chosen class. Two datasets with different degree of noise are used, where 10\% and 20\% of all the labels in the training dataset are reassigned.  

\begin{figure}[!htb]
    \centering
    \includegraphics[width=\textwidth]{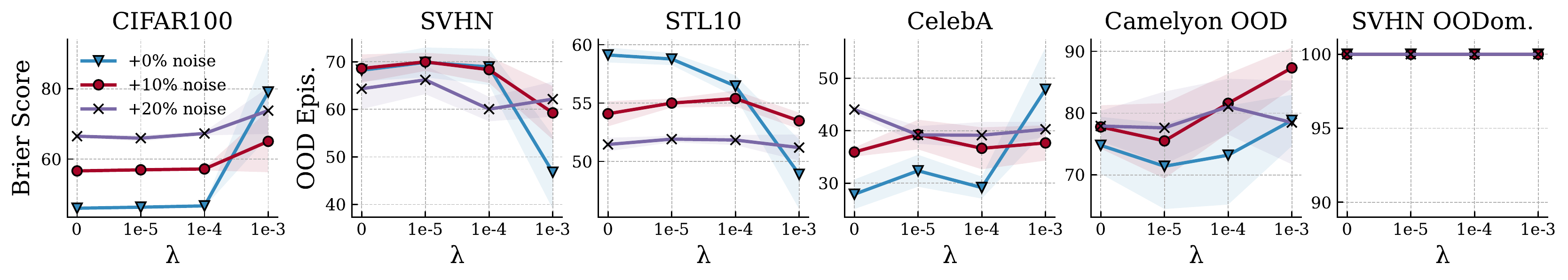}
    \includegraphics[width=\textwidth]{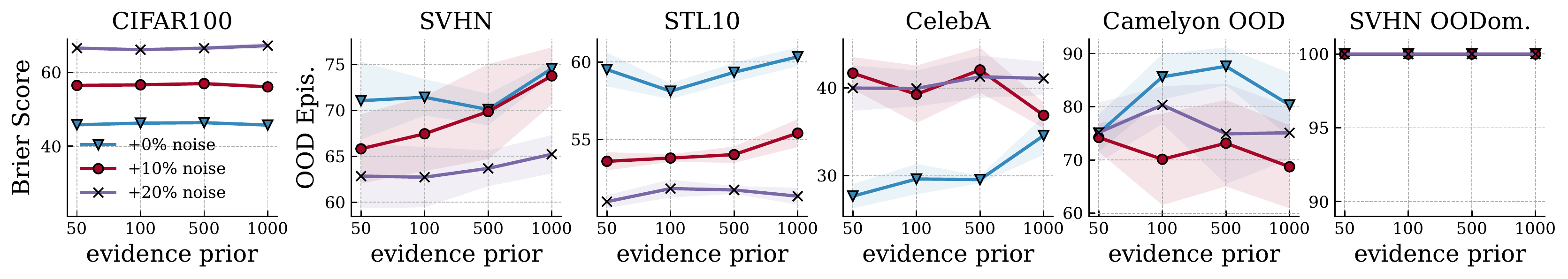}
    
    \caption{Results of enforcing different \textbf{prior} in NatPN on CIFAR100 by changing the (top) \textit{entropy regularization} $\lambda$ and the (bottom) \textit{evidence prior} $n^{prior}$. Different priors do not lead consistent results improvements.}
    \label{fig:prior_ood_brier}
\end{figure}

\end{document}